\documentclass[11pt]{article}

\PassOptionsToPackage{hyphens}{url}

\usepackage[final]{acl}

\usepackage{latexsym}

\usepackage[T1]{fontenc}

\usepackage[english,bidi=default]{babel} 
\babelfont{rm}{TeXGyreTermesX} 

\usepackage[utf8]{inputenc}

\usepackage{microtype}

\usepackage{amssymb}
\usepackage{fontspec}
\usepackage{polyglossia}

\setdefaultlanguage{english}
\setotherlanguage{thai}

\newfontfamily\thaifont[
    Path=./fonts/, Extension=.ttf,
    UprightFont=*-Regular, BoldFont=*-Bold,
    ItalicFont=*-Italic, BoldItalicFont=*-BoldItalic,
    Script=Thai,
    Language=Thai,
    Renderer=HarfBuzz
]{Sarabun}

\usepackage[most]{tcolorbox}
\usepackage{booktabs}
\usepackage{multirow}
\usepackage{makecell}
\usepackage{enumitem}

\usepackage{siunitx}
\usepackage{graphicx}
\usepackage{tabularx}
\usepackage{array}
\usepackage{pifont}

\newcommand{\cmark}{\ding{51}}
\newcommand{\xmark}{\ding{55}}
\newcommand{\thaitext}[1]{{\scriptsize\textthai{#1}}}
\newcommand{\thaicell}[1]{\begin{minipage}[t]{\linewidth}\raggedright
  \linespread{1.3}\scriptsize\lineskiplimit=-\maxdimen\XeTeXlinebreaklocale "th"\XeTeXlinebreakskip=0pt plus 0.1pt
  \textthai{#1}\end{minipage}}
\newcommand{\entext}[1]{{\scriptsize #1}}

\title{ASCRIBE: Atomic and Significance-Based Reasoning for Thai Clinical SOAP Note Generation}

\author{
 \textbf{Tarm Kalavantavanich\textsuperscript{1,*}},
 \textbf{Teerawut Ponarchar\textsuperscript{1,*}},
 \textbf{Pattaramanee Arsomngern\textsuperscript{1,*}},
\\
 \textbf{Jenta Wonglertsakul\textsuperscript{2,\(\dagger\)}},
 \textbf{Watcharakorn Chuthong\textsuperscript{1}},
 \textbf{Chiraphat Boonnag\textsuperscript{1}},
\\
 \textbf{Knot Pipatsrisawat\textsuperscript{1}},
 \textbf{Titipat Achakulvisut\textsuperscript{1,3}}
\\
\\
 \textsuperscript{1}Looloo Health, Thailand,
 \textsuperscript{2}Kasetsart University, Thailand,
 \textsuperscript{3}Mahidol University, Thailand
\\
 \small{
  \textsuperscript{*}Equal contribution.
 \textsuperscript{\(\dagger\)}Work conducted during an internship at Looloo Health.
 }\\
 \small{
\texttt{\{tarm.kalavan, teerawut.p, pattaramanee.a, c.boonnag, knot\}@loolootech.com}}  \\
\small{\texttt{jenta.w@live.ku.th} \qquad \texttt{watcharakorn.c@looloohealth.com} \qquad\texttt{titipat.ach@mahidol.ac.th}}
}

\usepackage{fontspec}
\newfontfamily\promptfont{TlwgTypewriter}[
  Path=./fonts/, Extension=.ttf,
  UprightFont=*, BoldFont=*-Bold,
  ItalicFont=*-Oblique, BoldItalicFont=*-BoldOblique, Scale=0.85]
\usepackage[most]{tcolorbox}
\usepackage{fvextra}   

\newtcolorbox{promptbox}[1]{enhanced, width=\textwidth, colback=gray!5,
  colframe=black, title={#1}, fonttitle=\bfseries, boxrule=0.5pt, sharp corners}
\newfontfamily\promptthaifont[
    Path=./fonts/, Extension=.ttf,
    UprightFont=*-Regular, BoldFont=*-Bold,
    ItalicFont=*-Italic, BoldItalicFont=*-BoldItalic,
    Script=Thai, Language=Thai, Renderer=HarfBuzz, Scale=0.85
]{Sarabun}
\newcommand{\promptthai}[1]{\textthai{\promptthaifont #1}}
\newcommand{\promptsize}{\fontsize{8}{9}\selectfont}

\newcommand{\ascribe}{\textsc{ASCRIBE}}

\begin{document}
\maketitle
\begin{abstract}
Automatic SOAP note generation can ease the documentation burden on physicians, but existing reasoning methods often omit clinically important information and generate unsupported content. Progress in Thai is further hindered by the lack of publicly available datasets.
We propose \ascribe{}, a physician-inspired reasoning framework that ascribes a clinical-significance level to each extracted atomic fact in the conversation before summarization, making a general-purpose LLM a more reliable scribe. We also release ThaiClinicBench, the first de-identified Thai clinical summarization benchmark of real encounters, together with a synthetic training corpus derived from real clinical notes.
As a prompt, ASCRIBE outperforms chain-of-thought prompting on GPT-5.4 and Gemini 3.1 Pro across the physician-aligned LLM-judge metrics and improves on standard prompting by up to 10.3 points on the completeness LLM-judge metric. As a GRPO reward, it enables a Gemma-4-E4B model trained solely on synthetic data to match Gemini 3.1 Pro in factual precision and surpass it in completeness. Code and data can be found at \url{https://github.com/loolootech/ascribe}.

\end{abstract}

\section{Introduction}

Comprehensive clinical documentation is critical for patient safety yet highly time-consuming, a serious burden in regions with low doctor-to-patient ratios such as Southeast Asia~\cite{kanchanachitra2011human}. AI medical scribes commonly address this through a transcribe-summarize pipeline~\cite{finley-etal-2018-automated}: an automatic speech recognition (ASR) system transcribes the encounter, and a summarization system converts it into a structured note such as the SOAP (Subjective, Objective, Assessment, Plan) format~\cite{weed1968medical}. However, transcripts are error-prone due to the spontaneous, disfluent nature of clinical conversation. This problem is compounded in low-resource languages like Thai, whose conversational and dialectal nuances challenge even frontier large language models (LLMs)~\cite{limkonchotiwat2025assessing}.


Prior work on clinical summarization primarily fine-tunes models on paired dialogue transcripts and summaries from English datasets~\cite{acibench,abacha2023empirical}. When such data are limited, recent efforts synthesize transcripts from patient profiles via agent simulation~\cite{jenDataSynth}. For prompting techniques, while chain-of-thought (CoT) has been used to encourage clinical reasoning across medical tasks~\cite{nachane2024fewShotCOT,wu2024guiding}, it often accumulates errors, resulting in poorly grounded outputs~\cite{whycotfails}. Crucially, these efforts remain unvalidated for Thai: to our knowledge, no public Thai dataset supporting SOAP note generation exists, leaving both data and reasoning gaps unresolved.


To close the data gap, we adapt the transcript synthesis pipeline of \citet{jenDataSynth}. By leveraging a real internal corpus of paired transcripts and SOAP notes, we synthesize a new Thai training corpus, S-ICOPD, which we release along with ThaiClinicBench, a de-identified benchmark of real encounters and physician-authored SOAP notes collected from a Thai clinic.

\begin{figure*}[t] 
    \centering
    \includegraphics[width=\textwidth]{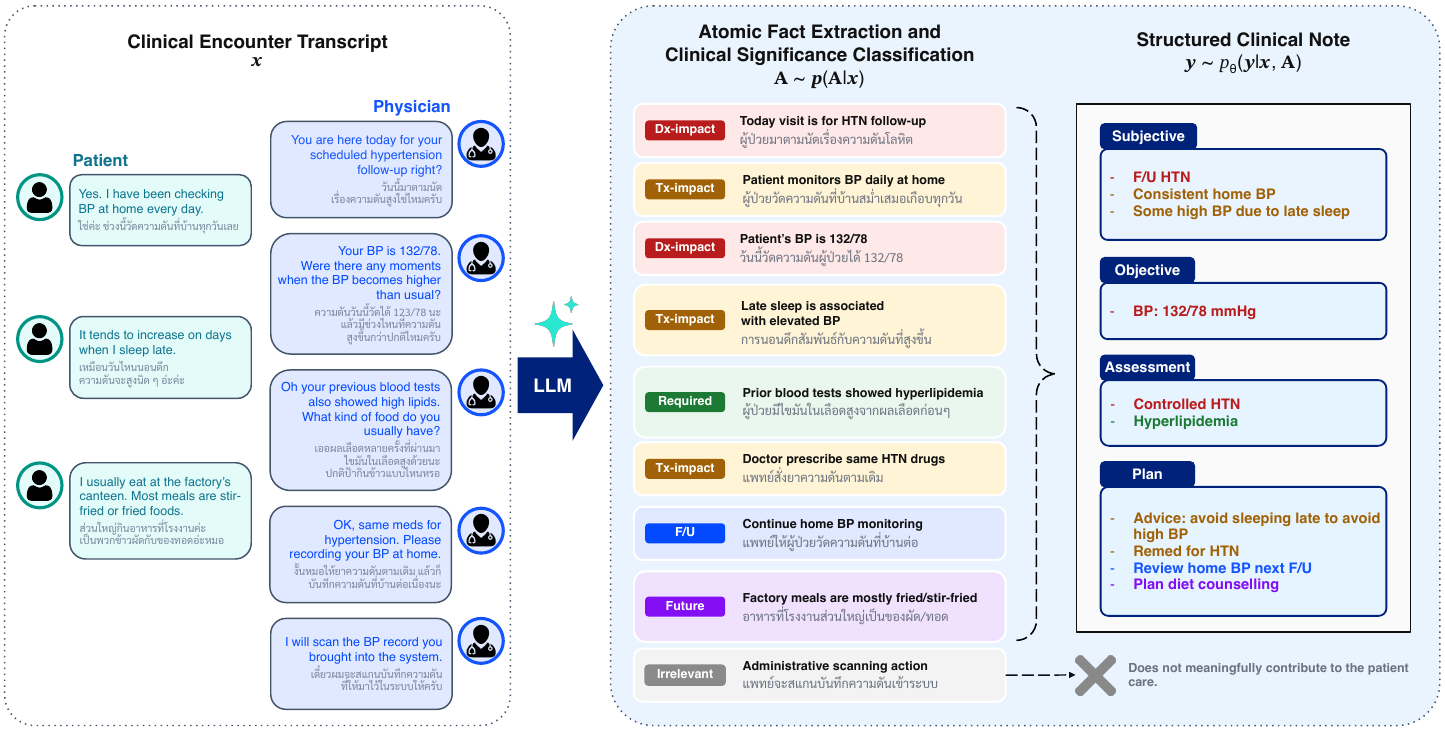} 
    \caption{\textbf{ASCRIBE's SOAP generation pipeline.} Atomic facts are extracted from the transcript $x$ and classified by their clinical significance to form the reasoning trace $A$, which conditions generation of the final SOAP note $y$.}
    \label{fig:atomic-medsig-diagram}
\end{figure*}

To close the reasoning gap, we introduce \ascribe{} (\textbf{A}tomic, \textbf{S}ignificance-Based \textbf{C}linical \textbf{R}easoning Framework for Medical Scr\textbf{IBE}), a clinically grounded reasoning framework that improves note quality both as a prompt template and as Group Relative Policy Optimization (GRPO)~\cite{GRPO} rewards. Inspired by physician history-taking, \ascribe{} decomposes a dialogue into atomic facts~\cite{factscore,densex}, ranks them by clinical significance, and generates the SOAP note from this reasoning trace (Figure~\ref{fig:atomic-medsig-diagram}). We evaluate notes on Thai encounters with two physician-aligned LLM-as-a-judge metrics, ClaimEntailment (completeness) and SOAPPrecision (factual grounding), that also serve as verifiable RL rewards. As a prompt, \ascribe{} improves frontier LLMs by up to 10.3 points over standard prompting. As GRPO rewards, \ascribe{} lets a fine-tuned Gemma-4-E4B~\cite{gemmateam2026gemma4technicalreport} surpass Gemini 3.1 Pro on completeness by over 10 points, even when trained only on synthetic data.

In summary, our contributions are:
\begin{itemize}[]
    \item \ascribe{}, a physician-inspired framework that ranks atomic clinical facts by significance and improves SOAP notes both as a prompt and as GRPO rewards.
    \item A Thai transcript-synthesis pipeline, with real-data-seeded synthetic transcripts matching real-transcript SFT performance and surpassing it on completeness under GRPO.
    \item Two released datasets: S-ICOPD, the synthetic training corpus, and ThaiClinicBench, the first medical-scribe benchmark from a Thai clinic.
\end{itemize}

\section{Related Work}
\paragraph{Clinical Note Generation.} 
Early work developed section-aware SOAP note generation pipelines~\citep{krishna2021generating}, with subsequent studies releasing English datasets and benchmarks~\citep{abacha2023empirical,abacha2023overview,acibench}. Later work improved performance using in-domain pre-training~\citep{grambow2022domain}, extractive-abstractive summarization~\citep{neupane2024clinicsum}, transcript synthesis~\citep{jenDataSynth}, and prompting techniques~\citep{kernberg2024chatgpt4}. Yet LLM-generated notes still suffer from omission, hallucination, and poor out-of-domain generalization~\citep{kernberg2024chatgpt4,chen2024exploring}. 


\paragraph{Reasoning in Medical Tasks.}
LLMs show strong medical reasoning using chain-of-thought (CoT) prompting~\cite{cot}, with Med-PaLM 2 reaching expert-level accuracy on medical licensing exams~\citep{singhal2025medpalm2}. Recent work grounds this reasoning in clinical knowledge, such as anchoring it in knowledge seeds~\citep{wu2024guiding} or mirroring physicians' diagnosis with few-shot examples~\citep{nachane2024fewShotCOT}. However, recent evidence shows that CoT often degrades performance on clinical text, with failures driven by hallucination and weak grounding rather than faithful reasoning~\citep{whycotfails}; this extends to SOAP note generation from clinical dialogue~\citep{Faisal2026WhenRH}. Motivated by these findings, we instead enforce structured reasoning over atomic clinical facts, not free-form generation.

\paragraph{Evaluating Clinical Notes and RLVR.}
Conventional overlap metrics such as ROUGE correlate inconsistently with physician judgment and miss clinically critical omissions and hallucinations~\citep{abacha2023investigation}. Instead, prior work evaluates notes by detecting omitted atomic facts~\citep{schumacher2023medomit}, grading errors by clinical impact~\citep{asgari2025framework}, or measuring claim-level precision and recall~\citep{doclens}, the last of which was adapted into a GRPO reward for reinforcement learning with verifiable rewards (RLVR)~\citep{GRPO,jhaveri2025optimizing}. 
Unlike prior work that rewards only the final note, we also reward the reasoning trace by supervising clinical fact extraction and significance attribution.


\section{Proposed Method}
We formulate SOAP note generation as a sequence-to-sequence summarization task. Given a transcript of a physician--patient encounter, a language model produces a structured clinical note in SOAP format, comprising the Subjective findings reported by the patient, the Objective findings from physical exams and diagnostic results, the physician's Assessment, and the management Plan. Concretely, we follow the SOAP-inspired four-division scheme of \citet{acibench} in our experiments, with each section emitted as a list of entries, which we refer to as \emph{statements}. 

We approach this task from two angles. First, we study LLM summarization. Since conventional prompting strategies such as CoT struggle to improve clinical note quality~\cite{whycotfails,Faisal2026WhenRH}, we propose a sequence of reasoning steps and a reasoning-driven fine-tuning method to address it. Second, we tackle clinical data scarcity in low-resource languages like Thai, by exploring synthetic transcript generation as a fine-tuning dataset and outlining how such encounter transcripts could be synthesized from a clinical note.





\subsection{\ascribe{}: Reasoning over Atomic Facts and Clinical Significance}
\label{sec:atomicmedsig}

Physicians conducting history-taking listen for key facts as they emerge, directing their attention towards clinically impactful statements while filtering out irrelevant ones. Inspired by this process, we introduce \ascribe{}, a structured reasoning framework that decomposes clinical summarization into two steps: atomic fact extraction and clinical significance classification, as illustrated by Figure~\ref{fig:atomic-medsig-diagram}.

\paragraph{Atomic Fact Extraction.}
To ground the model's reasoning in explicit, verifiable units of information, we first use an LLM to extract atomic facts from the transcript before summarization. Inspired by prior work on atomic facts and propositions~\cite{factscore,densex}, we define three conditions tailored to clinical text:
\begin{itemize}[]
    \item \emph{Standalone completeness}: Each fact must be a self-contained, natural sentence that is interpretable on its own, with all references resolved to explicit subject names.
    \item \emph{One idea}: Each fact must contain a single idea, with compound statements split into separate facts unless a clause is not meaningful on its own (e.g., a causal or temporal clause), in which case it stays attached so the main fact remains complete. 
    \item \emph{Uniqueness}: A fact stated multiple times must appear only once, in its most complete and specific form. 
\end{itemize} 

In our experiments, we extract facts with an LLM whose instruction was tuned against LLM judges, referred to as the \emph{Atomic Fact LLM Judges}, that validate generated atomic facts against a physician-curated gold set in a precision-recall fashion and achieve substantial physician agreement (Appendix~\ref{app:atomicfactextraction}). 


\paragraph{Clinical Significance Classification.}

\begin{table}[t]
\centering
\small
\begin{tabular}{p{0.25\linewidth}p{0.70\linewidth}}
\toprule
\textbf{Level} & \textbf{Description} \\
\midrule
Diagnosis-impacting &
Essential information supporting the diagnosis or assessment of the current visit (e.g., chief complaint, symptoms, clinical findings, disease progression, physical examination, signs, laboratory or imaging findings). \\ \\

Treatment-impacting &
Details regarding or influencing the management of the current visit (e.g., medications, procedures, follow-up plans, treatment recommendations). \\ \\

Required &
Patient information required by clinical standards, regardless of the current complaint (e.g., comorbidities, drug allergies, smoking/alcohol history). \\ \\

Follow-up-relevant &
Details not essential to the current visit but useful for subsequent follow-ups (e.g., advices given, referrals, treatment goals). \\ \\

Future-reference &
Details useful for future reference but unlikely to affect the current episode of care (e.g., irrelevant family history, social history, unrelated medical history). \\ \\

Irrelevant &
Does not meaningfully contribute to current or future clinical decision-making (e.g., redundant statements, vague complaints, hyper-specific context, administrative details, physician explanations). \\
\bottomrule
\end{tabular}

\caption{Clinical significance levels used for categorizing atomic facts. Levels are listed from most to least significant.}
\label{tab:significance}
\end{table}

We define six levels of clinical significance, ranked by how much a fact could impact patient safety and care as detailed in Table~\ref{tab:significance}.

We co-designed these levels with certified physicians, drawing on causality found in clinical settings and the Problem-Oriented Medical Record~\cite{weed1968medical}, which underlies SOAP and is mandated in Thailand's national medical-record standards~\cite{nhso_mra}.
The first three levels mirror what a SOAP note must support, namely the assessment, the plan, and mandated documentation, while the remaining levels extend this logic to longitudinal care.  
An LLM classifier reproduces the physicians' labels with substantial agreement (Appendix~\ref{app:clinical_sig}).

\paragraph{Structured Reasoning-based Generation.}
We combine the fact extraction and classification into a single reasoning trace. Given a transcript $x$, the model $p_\theta$ first generates the trace $A$ as a set of atomic facts with their evidence and clinical significance,
\begin{equation}
    A=\{(f_i,e_i,s_i)\}_{i=1}^{N},
\end{equation}
where $f_i$ is an atomic fact, $e_i$ is a short supporting span from the transcript that grounds it, $s_i$ is its predicted clinical significance level, and $N$ is the number of extracted atomic facts. 
The model then generates the SOAP note $y$ in the same decoding pass, conditioned on both the transcript and the reasoning trace,
\begin{equation}
y\sim p_{\theta}(y \mid x, A).
\end{equation}
Conditioning on $A$ encourages the model to identify and prioritize clinically impactful facts before producing the SOAP note, rather than summarizing directly or relying on free-form reasoning.


\subsection{Fine-Tuning Small Language Models for SOAP Note Summarization}

\begin{figure}[t] 
    \centering
    \includegraphics[width=\columnwidth]{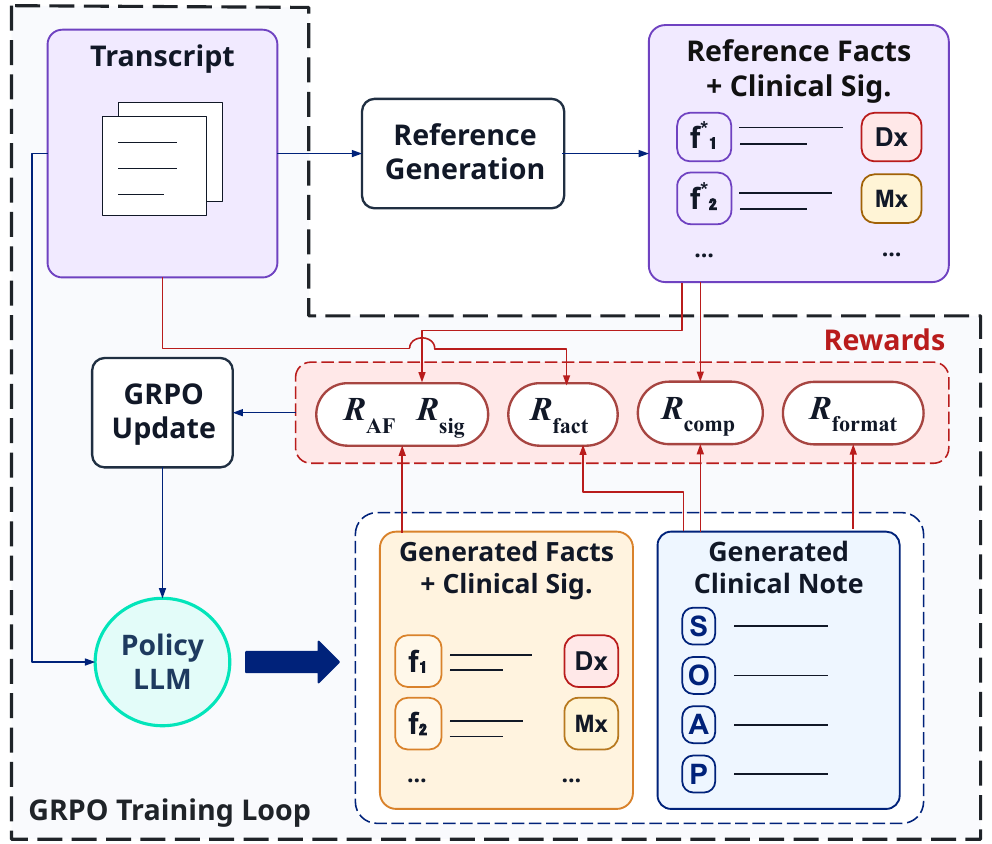} 
    \caption{\textbf{GRPO training pipeline.} The generated \ascribe{} reasoning trace $A$ and SOAP note from the policy LLM are scored by the five rewards of Section~\ref{sec:rlvr}, which jointly drive the GRPO update. Certain rewards require reference facts whose generation details are in Section~\ref{sec:reference_summary_gen}.}
    \label{fig:rlvr}
\end{figure}

Due to their capability to be deployed in on-premise settings, Small Language Models (SLMs) are our candidate summarization models, for which we explore both supervised and reinforcement learning techniques.

\subsubsection{Supervised Fine-Tuning (SFT)}
We apply LoRA-based SFT~\cite{lora} in two settings: standard prompting, which trains the model to generate only SOAP notes from clinical transcripts; and structured reasoning, where the target is prepended with a reference reasoning trace.

\subsubsection{Reinforcement Learning with Verifiable Rewards (RLVR)} 
\label{sec:rlvr}

To align the model with \ascribe{} reasoning, we adopt GRPO~\cite{GRPO} with LoRA as shown in Figure~\ref{fig:rlvr}. We use our structured reasoning to define five reward functions that optimize both reasoning and final note quality, each scaled to the range $[0, 1]$, as follows:

\paragraph{Atomic Fact Reward.}
Let $A$ and $A^*=\{(f_j^*,e_j^*,s_j^*)\}_{j=1}^{M}$ denote the sets of generated and reference atomic facts, respectively. We define two \emph{Atomic Fact LLM Judges} to compare the two sets independently and identify semantically matching facts (Appendix~\ref{app:overallatomic}). We denote $c(i)$ as the index of the reference fact $f_{c(i)}^*$ matched to the generated fact $f_i$ assigned by the Atomic Fact LLM Judge, with $c(i)=\varnothing$ when no match exists. Atomic precision and recall are computed as:

\[
\mathrm{Prec}_{\mathrm{atomic}}
=
\frac{\#\text{matched generated facts}}{N},
\]
\[
\mathrm{Rec}_{\mathrm{atomic}}
=
\frac{\#\text{matched reference facts}}{M}.
\]

The final reward is their harmonic mean:

\begin{equation}
R_{\mathrm{AF}}
=
\frac{
2\,
\mathrm{Prec}_{\mathrm{atomic}}
\,
\mathrm{Rec}_{\mathrm{atomic}}
}{
\mathrm{Prec}_{\mathrm{atomic}}
+
\mathrm{Rec}_{\mathrm{atomic}}
}.
\label{eq:atomicfact}
\end{equation}

\paragraph{Clinical Significance Reward.}
For each generated fact $f_i$ and its matched reference fact $f_{c(i)}^*$, we compare $s_i$ against the reference significance label $s_{c(i)}^*$. If no reference fact is matched, the prediction is considered incorrect. The reward is computed as:

\begin{equation}
    R_{\mathrm{sig}}=\frac{1}{N}\sum_{i=1}^{N}\mathbf{1}\!\left[s_i=s_{c(i)}^*\right].
\label{eq:clinicalsig}
\end{equation}

\paragraph{Format Reward.} $R_{\mathrm{format}}$ checks that the generated note follows the required SOAP note structure; the reward is reduced if a generated note does not follow the format.

\paragraph{Completeness Reward.}
Following evaluation-driven GRPO for clinical summarization~\cite{jhaveri2025optimizing}, we use the ClaimEntailment metric described in Section~\ref{sec:metric} as the $R_{\mathrm{comp}}$ reward. This directly encourages generated SOAP notes to preserve clinically important facts from the encounter.

\paragraph{Factual Grounding Reward.}
Following \citet{jhaveri2025optimizing}, we use the SOAPPrecision metric outlined in Section~\ref{sec:metric} as the $R_{\mathrm{fact}}$ reward, which encourages clinical statements in the SOAP note $y$ to remain factually supported by the encounter transcript.


\subsection{Synthesizing Encounter Transcripts}
\label{sec:synth}

As real clinical transcripts are scarce in low-resource languages like Thai, we explore synthetic transcript generation by adapting the pipeline of \citet{jenDataSynth} to our settings. Our pipeline takes a SOAP note as the seed patient profile, from which the physician--patient dialogue transcripts were synthesized in two stages. First, \emph{intent planning} constructs a weighted directed intent graph that models probabilistic conversation flow and selects an intent path grounded in the profile. Second, \emph{transcript generation} synthesizes the dialogue based on that path. See Appendix~\ref{app:synth} for full details.

\paragraph{Intent Planning.}
Following \citet{jenDataSynth}, we construct an intent graph from a corpus of real physician--patient conversations. Each training transcript is first abstracted into an \emph{intent skeleton}, a sequence of conversational intents discussed during the encounter. These intent skeletons are then aggregated into an intent graph $G = (V, E)$, where $V$ represents intents and $E$ represents transitions whose weights refer to the transition probabilities.
%
%
For each seed patient profile, we gather candidate intent paths by performing depth-first search on $G$. To ensure the path is consistent with the profile, we limit the search space to intent nodes corresponding to the profile, then rank the resulting paths with an LLM judge, selecting the one with the highest score.


\paragraph{Transcript Generation.}
Given a patient profile and the chosen intent path, a doctor agent and a patient agent generate the encounter transcript, addressing one intent at a time. Generation for each intent continues until an LLM judge determines that the intent has been fulfilled or the per-intent turn limit has been reached. Each completed dialogue section is then validated against ten rules adapted from \citet{jenDataSynth} for Thai clinical conversations. If validation fails, the section is regenerated using the validator's feedback. The pipeline proceeds to the next intent once the section passes validation or the retry budget is exhausted, retaining the final attempt.

\section{Datasets}

We use three datasets in this study: an internal outpatient corpus of Thai physician--patient encounters (ICOPD), a new real-world benchmark (ThaiClinicBench), and a synthetic training corpus (S-ICOPD) built from ICOPD. All real conversation transcripts are manually produced by human annotators. We describe each dataset and how the reference training targets are constructed.

\subsection{Internal Clinical OPD Dataset (ICOPD)}
The ICOPD dataset is a private corpus of 400 consented outpatient-department (OPD) encounters collected in Thailand. The conversations are in colloquial Thai, with dialectal medical terms common in Thai clinical settings. For each encounter, we obtain the transcript, the physician-written note, and an LLM-constructed reference SOAP note for SFT (Section~\ref{sec:reference_summary_gen}).  
The conversations average 328 words (1.34k characters). We split the corpus into training (300), validation (50), and test (50) sets. 

\subsection{ThaiClinicBench}
\label{sec:clinic_dataset}
We curate ThaiClinicBench, a benchmark of manually de-identified and consented OPD encounters collected from a medical clinic in Thailand. It contains 44 encounters (pairs of transcript and physician-written SOAP note). The conversations average 703 words (2.89k characters). All encounters serve as an out-of-distribution test set. To the best of our knowledge, this is the first Thai medical-scribe benchmark released for research.
    
\subsection{Synthetic ICOPD Dataset (S-ICOPD)}
Beyond the real encounters, we construct a synthetic ICOPD using the transcript synthesis pipeline (Section~\ref{sec:synth}) as an additional training dataset. We seed it with the physician-written SOAP notes of ICOPD, each serving as the patient profile from which we synthesize an encounter transcript. Running it on the training and validation splits yields 295 and 50 synthetic transcripts, respectively, with 5 training cases excluded as the pipeline does not produce a usable transcript. Each synthetic transcript is then re-summarized with the pipeline of Section~\ref{sec:reference_summary_gen} to obtain its reference data, so that every text in S-ICOPD is model-generated.

\subsection{Reference Training Target Generation}
\label{sec:reference_summary_gen}

We generate a reference set of atomic facts and their significance labels for each encounter using Gemini 2.5 Pro with prompts in Appendix~\ref{app:overallatomic}. These serve as the reference facts behind $R_\mathrm{AF}$ and $R_\mathrm{sig}$ during RLVR, the input to evaluation metrics in Section~\ref{sec:metric}, and the reasoning targets for SFT.

For SFT, we construct reference SOAP notes in two stages. We first generate a draft SOAP note from the transcript alone using Gemini 3.1 Pro conditioned on the proposed structured reasoning and then refine it with GPT-5.4 (Appendix~\ref{app:refineprompt}) to further improve overall quality.


\section{Evaluation Metrics}

\label{sec:metric}



Following \citet{acibench}, we calculate ROUGE-L~\cite{lin2004rouge} and BLEURT~\cite{sellam2020bleurt} to measure lexical and semantic similarity between generated notes and SOAP notes written by certified physicians. To better assess clinical quality, we adopt an LLM-judge framework inspired by DocLens~\cite{doclens}, simplifying their Claim Precision pipeline by judging the generated note directly rather than extracting claims from it first. These adapted LLM judges compute two complementary metrics and are validated for alignment with physicians (details are provided in Appendix~\ref{app:llmjudge}).

\paragraph{ClaimEntailment (CE)} measures completeness of the generated SOAP note by using reference atomic facts to evaluate whether each fact was entailed by the generated note (i.e., recall). We also introduce \emph{CE-Impact}, which calculates CE only on the three highest clinical significance levels (\emph{Diagnosis-impacting}, \emph{Treatment-impacting}, and \emph{Required}) to further measure the recall of clinically important facts.

\paragraph{SOAPPrecision (SP)} measures factual grounding by verifying whether each statement in the generated SOAP note is explicitly supported by the encounter transcript (i.e., precision). It allows for paraphrasing and terminology normalization (e.g., from lay language to medical jargon) but penalizes unsupported, inferred, or hallucinated content.

\section{Experimental Settings}
\paragraph{Prompting Baselines.}
We use GPT-5.4 and Gemini 3.1 Pro as our baselines under three zero-shot prompting strategies: standard SOAP note generation~\cite{acibench}, CoT, and our proposed structured reasoning (see Appendix~\ref{app:prompt} for prompt details).

\paragraph{Fine-tuning Baselines.} We use Gemma-4-E4B-it~\cite{gemmateam2026gemma4technicalreport} as the base SLM for all fine-tuning experiments. We compare our solution against three GRPO baselines:
\textbf{ROUGE-L} uses the ROUGE-L score between the generated and reference SOAP notes as the reward, representing a conventional lexical-overlap optimization baseline~\cite{serl}.
\textbf{BGE-M3} uses BGE-M3~\cite{bgem3} embedding similarity between the generated and reference SOAP notes as the reward, following prior work~\cite{thailaw} that adopts it for fine-tuning Thai LLMs.
\textbf{ClinicEval} uses Completeness and Factual Grounding rewards following the evaluation-integrated GRPO training for clinical summarization~\cite{jhaveri2025optimizing}.
See Appendix~\ref{app:trainings} for training details.




\section{Experimental Results}
\begin{table*}[t]
\centering

\setlength{\tabcolsep}{2.6pt}
\renewcommand{\arraystretch}{1.12}

\resizebox{\textwidth}{!}{%
\begin{tabular}{llllccccc|ccccc}
\toprule
& & & &
\multicolumn{5}{c}{\textbf{ICOPD}} &
\multicolumn{5}{c}{\textbf{ThaiClinicBench}} \\
\cmidrule(lr){5-9}
\cmidrule(lr){10-14}

\makecell{Model /\\Training} &
\makecell{GRPO\\Reward} &
\makecell{Training\\Data} &
Prompt &
SP &
CE &
\makecell{CE-\\Impact} &
ROUGE-L &
BLEURT &
SP &
CE &
\makecell{CE-\\Impact} &
ROUGE-L &
BLEURT \\
\midrule

\multicolumn{14}{c}{\textit{Frontier LLM prompting}} \\
\midrule

\multirow{3}{*}{Gemini 3.1 Pro}
& -- & -- & Standard
& 83.5 & 52.6 & 63.3 & \underline{\textbf{0.3467}} & \underline{\textbf{0.3366}}
& 86.6 & 51.6 & 58.8 & \underline{\textbf{0.3390}} & 0.2488 \\

& -- & -- & CoT
& 81.0 & 53.2 & 64.3 & 0.3387 & 0.3177
& 84.7 & 52.0 & 59.8 & 0.3205 & 0.2430 \\

& -- & -- & \textbf{Ours}
& \underline{86.9} & \underline{54.8} & \underline{67.0} & 0.3406 & 0.3220
& \underline{89.2} & \underline{58.4} & \underline{67.9} & 0.3170 & \underline{0.2525} \\

\cmidrule(lr){1-14}

\multirow{3}{*}{GPT-5.4}
& -- & -- & Standard
& 90.5 & 54.5 & 64.0 & \underline{0.3413} & \underline{0.3299}
& 92.5 & 55.7 & 64.3 & \underline{0.3023} & 0.2623 \\

& -- & -- & CoT
& 90.9 & 58.6 & 68.2 & 0.3169 & 0.3194
& 93.3 & 58.6 & 68.8 & 0.2809 & 0.2545 \\

& -- & -- & \textbf{Ours}
& \underline{\textbf{95.2}} & \underline{\textbf{59.6}} & \underline{\textbf{71.2}} & 0.3032 & 0.3237
& \underline{\textbf{98.3}} & \underline{\textbf{62.6}} & \underline{\textbf{74.6}} & 0.2781 & \underline{\textbf{0.2627}} \\

\midrule
\multicolumn{14}{c}{\textit{Gemma-4-E4B-it fine-tuning}} \\
\midrule

\multirow{2}{*}{Zero-shot}
& -- & -- & Standard
& 84.7 & 38.4 & 46.6 & \underline{0.3180} & 0.3183
& \underline{87.7} & 42.5 & 48.0 & \underline{0.3003} & 0.2707 \\

& -- & -- & \textbf{Ours}
& \underline{87.0} & \underline{44.2} & \underline{53.4} & 0.3026 & \underline{0.3207}
& 87.0 & \underline{46.6} & \underline{52.9} & 0.2915 & \underline{\textbf{0.2803}} \\

\cmidrule(lr){1-14}

\multirow{6}{*}{SFT}
& -- & Real & Standard
& 75.1 & \underline{46.6} & \underline{57.3} & 0.3420 & \underline{0.3224}
& 83.9 & 50.6 & 56.5 & \underline{0.3083} & \underline{0.2593} \\

& -- & Real & \textbf{Ours}
& 79.3 & 46.3 & 57.0 & 0.3395 & 0.2946
& 85.3 & \underline{53.8} & \underline{59.5} & 0.2961 & 0.2545 \\

& -- & Synth & Standard
& \underline{80.9} & 43.4 & 53.4 & 0.3272 & 0.3159
& 83.7 & 49.1 & 56.1 & 0.2928 & 0.2506 \\

& -- & Synth & \textbf{Ours}
& 77.2 & 45.5 & 55.7 & 0.3276 & 0.3060
& 82.9 & 50.8 & 57.0 & 0.2863 & 0.2520 \\

& -- & Real+Synth & Standard
& 78.5 & 41.9 & 51.1 & 0.3433 & 0.3112
& 83.0 & 48.8 & 56.7 & 0.3050 & 0.2550 \\

& -- & Real+Synth & \textbf{Ours}
& 77.5 & 45.5 & 55.7 & \underline{\textbf{0.3491}} & 0.3164
& \underline{86.7} & 51.8 & \underline{59.5} & 0.2993 & 0.2483 \\

\cmidrule(lr){1-14}

\multirow{7}{*}{GRPO}
& ROUGE-L & Real & Standard
& 86.8 & 33.9 & 40.2 & \underline{0.3489} & 0.3034
& 87.9 & 35.7 & 40.3 & \underline{\textbf{0.3149}} & \underline{0.2725} \\

& BGE-M3 & Real & Standard
& 81.4 & 48.8 & 59.3 & 0.3120 & \underline{\textbf{0.3366}}
& 84.4 & 52.3 & 58.8 & 0.2946 & 0.2687 \\

& ClinicEval & Real & Standard
& 83.5 & 61.4 & 70.3 & 0.2293 & 0.3143
& 86.9 & 63.7 & 72.3 & 0.2052 & 0.2396 \\

& ClinicEval & Real & \textbf{Ours}
& 84.8 & 61.8 & 71.0 & 0.2551 & 0.2979
& 88.3 & 63.5 & 71.8 & 0.2323 & 0.2467 \\

& Ours & Real & \textbf{Ours}
& \underline{\textbf{87.1}} & 62.1 & 71.2 & 0.2773 & 0.2898
& 89.4 & 64.0 & 72.1 & 0.2487 & 0.2450 \\

& Ours & Synth & \textbf{Ours}
& 86.0 & \underline{\textbf{64.9}} & \underline{\textbf{75.7}} & 0.2478 & 0.2786
& \underline{\textbf{89.6}} & 69.7 & \underline{\textbf{77.9}} & 0.2143 & 0.2280 \\

& Ours & Real+Synth & \textbf{Ours}
& 83.2 & 64.4 & 75.1 & 0.2306 & 0.3203
& 85.2 & \underline{\textbf{70.0}} & 77.0 & 0.2105 & 0.2509 \\

\bottomrule
\end{tabular}%
}
\caption{
Comparison of ASCRIBE prompting and fine-tuning against baseline methods on the ICOPD test split and ThaiClinicBench. \emph{Ours} denotes the corresponding ASCRIBE prompt or reward, \emph{Real} denotes training on ICOPD, and \emph{Synth} denotes training on S-ICOPD. Higher values indicate better performance for all metrics. \underline{Underline} marks the best result within each model/training block, and \textbf{bold} marks the best result within each approach (frontier LLM prompting vs.\ Gemma-4-E4B-it fine-tuning). SP, CE, and CE-Impact are reported on a 0--100 scale.
}
\label{tab:main_results}
\end{table*}

\begin{figure*}[t]
    \centering
    \includegraphics[width=440pt]{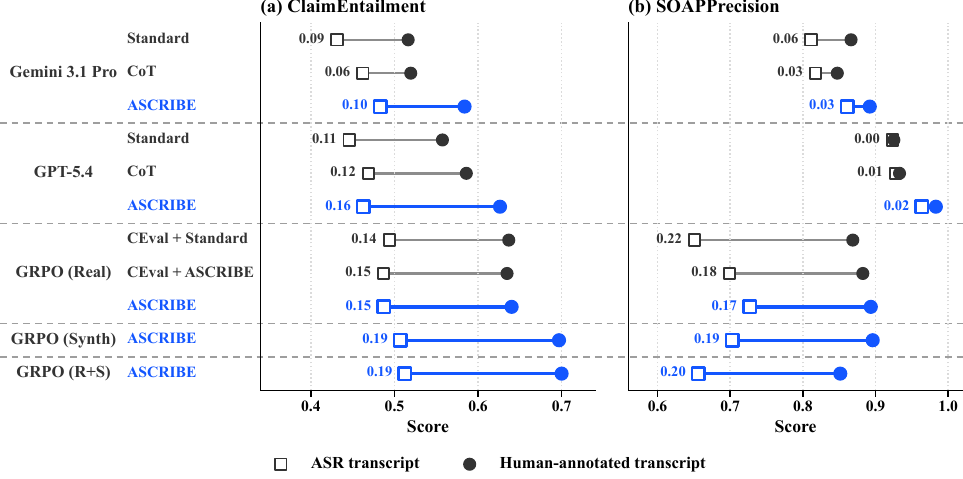}
    \caption{
    \textbf{Human-annotated vs.\ ASR transcript performance on ThaiClinicBench for CE and SP metrics.} \emph{R+S} refers to mixing real and synthetic training data and \emph{CEval} refers to the ClinicEval GRPO reward. Blue indicates settings that use the ASCRIBE prompt and, where applicable, the ASCRIBE GRPO reward. Labels denote the score drop from human-annotated to ASR transcripts. Our structured reasoning and synthetic training achieve the highest scores on ASR in several settings despite larger drops from human transcripts.}
    \label{fig:asr}
\end{figure*}

We evaluate the effectiveness of ASCRIBE for zero-shot prompting and SLM fine-tuning, compare SFT with GRPO, and analyze the contribution of the proposed reward. We further study the effectiveness of our synthetic training data, the relationship between conventional and clinical metrics, and robustness to ASR noise. Table~\ref{tab:main_results} outlines our full experimental results.

\paragraph{Effectiveness of Structured Reasoning.}
The \ascribe{} prompt outperforms both standard and CoT prompting in the zero-shot setting. For Gemini 3.1 Pro, the \ascribe{} prompt improves SP/CE/CE-Impact over standard prompting by +3.4/+2.2/+3.7 on ICOPD and +2.6/+6.8/+9.1 on ThaiClinicBench. GPT-5.4 shows even larger gains of +4.7/+5.1/+7.2 and +5.8/+6.9/+10.3, respectively. The same trend largely holds for Gemma-4-E4B in the zero-shot setting. In contrast, CoT prompting yields smaller CE gains while reducing conventional metrics and, in some cases, SP. For SFT settings, structured reasoning generally improves CE and CE-Impact, particularly on ThaiClinicBench, but its effect on SP is mixed.

\paragraph{SFT vs.\ GRPO.}
SFT generally achieves higher ROUGE-L, which could be attributed to being supervised against curated reference SOAP notes, but the differences in BLEURT scores are less consistent. In contrast, training models with GRPO using clinically grounded rewards substantially improves SP, CE, and CE-Impact, reflecting that lexical similarity alone does not necessarily represent the clinical quality of the notes.

\paragraph{Effectiveness of RLVR.}
Among GRPO variants trained on real ICOPD data, the ASCRIBE reward achieves the best overall clinical performance, obtaining the highest SP, CE, and CE-Impact on ICOPD (87.1/62.1/71.2) and similarly strong results on ThaiClinicBench (89.4/64.0/72.1). Trained with the ASCRIBE reward, Gemma-4-E4B also outperforms proprietary Gemini 3.1 Pro across all LLM-judge metrics and GPT-5.4 on CE, despite having only 4.5B effective parameters.

Notably, baseline performance varies substantially with the reward design. Optimizing for ROUGE-L maintains high SP but severely reduces CE, while optimizing for BGE-M3 improves CE but obtains the lowest SP of any GRPO baseline. Crucially, optimizing for ClinicEval substantially improves CE while maintaining comparable level of SP, demonstrating the value of directly optimizing for clinical quality. However, by rewarding the intermediate reasoning trace as well, training GRPO with \ascribe{} rewards improves both SP and CE.

\paragraph{Ablation of Structured Reasoning Prompt and Rewards.}
Focusing on the results of models trained on real data with GRPO reward, we found that the \ascribe{} prompt primarily improves SP, whereas our reward provides more consistent gains across the clinical metrics.
Substituting the standard prompt with the \ascribe{} prompt improves SP by +1.3 on ICOPD and +1.4 on ThaiClinicBench, while CE remains comparable. Under the \ascribe{} prompt, replacing the ClinicEval reward with our reward further improves SP/CE/CE-Impact by +2.3/+0.3/+0.2 on ICOPD and +1.1/+0.5/+0.3 on ThaiClinicBench.

\paragraph{Effectiveness of Synthetic Training Data.} 
Under SFT, training on synthetic data alone performs comparably to training on real data; on ThaiClinicBench, synthetic-only training with the standard prompt achieves SP/CE/CE-Impact of 83.7/49.1/56.1 versus 83.9/50.6/56.5. However, combining real and synthetic data lowers CE in every SFT settings, e.g., from 53.8 to 51.8 on ThaiClinicBench with our prompt despite a higher SP. These results suggest that synthetic data could substitute for real data in SFT, but naively mixing the two datasets does not bring further gains.

Under GRPO, synthetic-only training outperforms real-only training in CE and CE-Impact, improving from 62.1/71.2 to 64.9/75.7 on ICOPD and from 64.0/72.1 to 69.7/77.9 on ThaiClinicBench, with comparable SP. Mixed training yields similar CE, including the highest CE of 70.0 on ThaiClinicBench, but lower SP. These gains suggest that synthetic transcripts are also effective for RLVR.

We hypothesize that the performance degradation found in both SFT and GRPO settings stems from distributional differences between the corpora which were naively mixed. Real conversations contain disfluencies and colloquial expressions that might only be partially captured by synthetic transcripts.



\paragraph{Clinical vs.\ Conventional Evaluation.}
Frontier LLMs achieve strong clinical performance while also attaining competitive ROUGE-L and BLEURT scores. However, the trends across baselines show that conventional metrics do not consistently reflect clinical quality. For example, SFT and ROUGE-L-optimized GRPO can obtain high lexical-similarity scores without corresponding improvements in CE. Conversely, ClinicEval GRPO substantially improves CE and CE-Impact at the cost of ROUGE-L. These results suggest that conventional and clinical metrics capture complementary aspects of quality and that lexical similarity alone is insufficient for evaluating clinical notes.

\paragraph{Robustness to ASR Transcripts.}

Because real-world medical scribes rely on ASR systems, we evaluate robustness by replacing the human-annotated transcripts in ThaiClinicBench with ASR-generated transcripts (Appendix~\ref{app:asr}). Figure~\ref{fig:asr} compares model performance on human-annotated and ASR transcripts.
All models obtain lower CE and SP on ASR transcripts, confirming that transcription errors degrade note quality. Among frontier LLMs, our structured prompt achieves the highest ASR SP and improves CE over standard prompting for Gemini while remaining competitive for GPT-5.4.
For GRPO models, our reward improves ASR SP over ClinicEval (72.7 vs.\ 65.1--69.9), while models with synthetic data achieve the highest ASR CE. Overall, ASCRIBE variants improve absolute ASR performance in most settings, particularly in SP.
Nevertheless, GRPO models generally experience larger performance drops from human to ASR transcripts than frontier LLMs, highlighting the need for more robust training against ASR noise.

\section{Conclusion}
We presented ASCRIBE, a physician-inspired structured reasoning framework for Thai SOAP note generation. By explicitly modeling atomic clinical facts and their clinical significance, ASCRIBE prompting consistently outperforms standard and CoT prompting on frontier LLMs. As a training paradigm, ASCRIBE also provides the most effective reward formulation for GRPO fine-tuning, yielding the largest improvements in factual grounding. With our synthetic clinical encounters seeded from real clinical notes, ASCRIBE training achieves competitive performance without using additional real clinical conversations for model tuning, making the framework particularly attractive for clinical environments where data governance constrains access to external LLMs and real patient data. Finally, we release the synthetic clinical encounters and ThaiClinicBench, providing a practical foundation for developing clinically reliable medical scribes for Thai and other low-resource languages.

\section*{Limitations}
Our study is limited to a small number of outpatient encounters, a single SLM, and only the Thai language. This also includes the LLM judges that were used in our study. While they were validated for physician alignment and obtained a high level of alignment, they were validated on a limited number of encounters against a single board-certified physician, meaning inter-rater agreement could not be reported.
Although synthetic transcripts showed signs of improving the models on which they were trained, they may not fully capture the diversity and complexity of real clinical conversations yet. Even though the synthesis pipeline accepts SOAP notes from any source, including random generations, we used physician-written ones as the seed patient profile, so it is not fully synthetic. However, we re-summarized the synthetic encounters to ensure no leakage. We have not yet identified an optimal strategy for combining real and synthetic data, as naive mixing yields inconsistent improvements.
We also acknowledge that our primary metrics, namely ClaimEntailment and SOAPPrecision, also serve as GRPO rewards, so models trained with them are evaluated in part on their own training signal. Real user studies with physicians on the generated results are still lacking.
ASR errors still cause considerable performance degradation, particularly for GRPO-trained models, highlighting the need for more robust approaches to handle noisy transcripts. 
For ethical considerations, please refer to Appendix~\ref{app:ethics}. 


\section*{Acknowledgements}

This research has received funding support from the NSRF via the Research and Innovation Acceleration Agency for Competitiveness and Area Development (RCAD) (Program Management Unit for Frontier Brainpower and Future Industries) [grant number B13F690044].

\bibliography{custom}

\clearpage
\appendix
\section*{Appendix}
\label{sec:appendix}
This appendix provides supplementary methodological, implementation, and evaluation details for ASCRIBE. Note that the prompt figures show the system instructions only; the complete prompts and the input templates are available in our repository.
\begin{itemize}[leftmargin=*, nosep]
\item Appendix~\ref{app:prompt} presents the prompts for standard, chain-of-thought, and ASCRIBE-based SOAP note generation, together with the reference SOAP refinement prompt.

\item Appendix~\ref{app:overallatomic} describes the construction and validation of the reference atomic facts and clinical significance.

\item Appendix~\ref{app:synth} details the synthetic transcript generation pipeline.

\item Appendix~\ref{app:llmjudge} shows the ClaimEntailment and SOAPPrecision judges, with their alignment with physician annotations.

\item Appendices~\ref{app:trainings} and~\ref{app:asr} provide the SFT and GRPO training hyperparameters and the ASR robustness setup.

\item Appendix~\ref{app:ethics} details ethical considerations for each dataset.

\item Appendix~\ref{app:qualitative} presents qualitative examples of factual grounding and clinical information coverage.

\end{itemize}

\section{SOAP Note Generation Prompt}
\label{app:prompt}

This section provides the prompts used throughout the experiments, for both frontier LLMs and fine-tuned SLMs. All prompting experiments were conducted with the temperature set to 0.

Across all prompts, the note is returned as JSON with each of the four SOAP sections given as a list of statements, sometimes referred to as \texttt{points} in the prompts.

\subsection{Standard Prompt}
\label{app:defaultprompt}
\begin{figure*}[t]
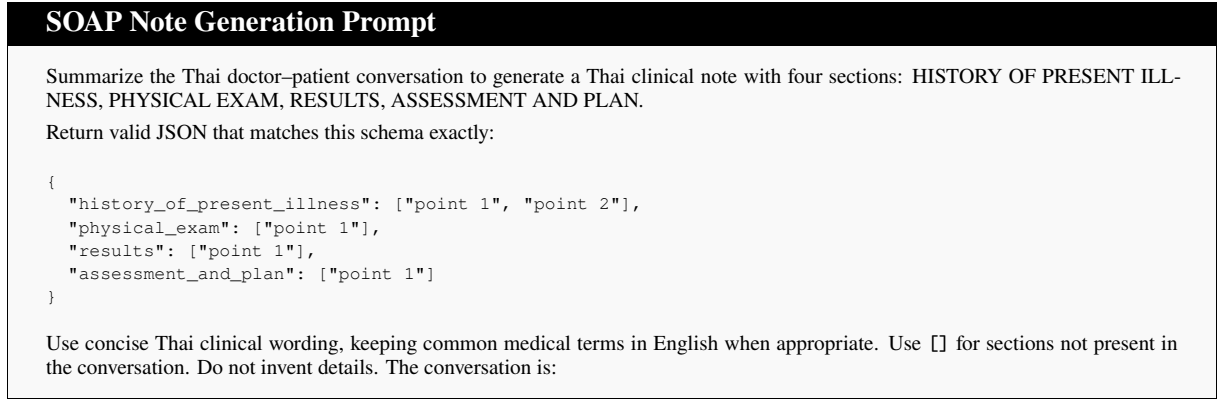

\centering
\begin{promptbox}{SOAP Note Generation Prompt}
\promptsize
Summarize the Thai doctor--patient conversation to generate a Thai clinical note with four sections: HISTORY OF PRESENT ILLNESS, PHYSICAL EXAM, RESULTS, ASSESSMENT AND PLAN.

\smallskip
Return valid JSON that matches this schema exactly:
\begin{Verbatim}[formatcom=\promptfont, fontsize=\promptsize]
{
  "history_of_present_illness": ["point 1", "point 2"],
  "physical_exam": ["point 1"],
  "results": ["point 1"],
  "assessment_and_plan": ["point 1"]
}
\end{Verbatim}

Use concise Thai clinical wording, keeping common medical terms in English when appropriate. Use \texttt{[]} for sections not present in the conversation. Do not invent details. The conversation is:
\end{promptbox}
\caption{Standard prompt used for SOAP note generation following the ACI-BENCH~\cite{acibench} study.}
\label{fig:defaultprompt}
\end{figure*}

We adopt the standard SOAP note generation prompt from ACI-BENCH~\cite{acibench}. Given a physician--patient conversation, the model is instructed to generate a structured SOAP note in JSON format containing the four sections: Subjective's \textit{History of Present Illness}, Objective's \textit{Physical Exam}, Objective's observable and diagnostic \textit{Results}, and \textit{Assessment and Plan}. The complete prompt is shown in Figure~\ref{fig:defaultprompt}.

\subsection{Chain-of-Thought Prompt}
Following \citet{whycotfails}, we extend the standard prompt in Appendix~\ref{app:defaultprompt} by requiring the model to produce an explicit chain-of-thought reasoning process before generating the final SOAP note. The output additionally contains a free-form \texttt{reasoning\_steps} field while keeping the same SOAP note generation objective. The complete prompt is shown in Figure~\ref{fig:cotprompt}.
\begin{figure*}[t]
\centering
\begin{promptbox}{SOAP Note Generation Prompt with Chain-of-Thought}
\promptsize
Summarize the Thai doctor--patient conversation to generate a Thai clinical note with four sections: HISTORY OF PRESENT ILLNESS, PHYSICAL EXAM, RESULTS, ASSESSMENT AND PLAN.

\smallskip
Solve it in a step-by-step fashion, return your answer in the following format, PROVIDE DETAILED ANALYSIS BEFORE THE RESULT:

\smallskip
Return valid JSON that matches this schema exactly:
\begin{Verbatim}[formatcom=\promptfont, fontsize=\promptsize]
{
  "reasoning_steps": "free-form analysis of how the conversation maps to each SOAP section",
  "history_of_present_illness": ["point 1", "point 2"],
  "physical_exam": ["point 1"],
  "results": ["point 1"],
  "assessment_and_plan": ["point 1"]
}
\end{Verbatim}

Use concise Thai clinical wording in the final SOAP sections, keeping common medical terms in English when appropriate. Use \texttt{[]} for sections not present in the conversation. Do not invent details. The conversation is:
\end{promptbox}
\caption{CoT for SOAP note generation based on ACI-BENCH~\cite{acibench} and \citet{whycotfails}.}
\label{fig:cotprompt}
\end{figure*}

\subsection{\ascribe{} Structured Reasoning Prompt}
\label{app:e2eprompt}
We propose \ascribe{} prompt, our physician-inspired structured reasoning prompt, in Section~\ref{sec:atomicmedsig}. It first decomposes the conversation into atomic clinical facts, predicts the clinical significance of each fact, and then generates the final SOAP note conditioned on the resulting reasoning trace. Unlike free-form chain-of-thought prompting, the intermediate reasoning is explicitly structured and grounded in transcript evidence. The complete prompt is shown in Figure~\ref{fig:e2eprompt}.

\begin{figure*}[t]
\centering
\begin{promptbox}{SOAP Note Generation Prompt with Structured Reasoning}
\promptsize
Summarize the Thai doctor--patient conversation to generate a Thai clinical note with four sections: HISTORY OF PRESENT ILLNESS, PHYSICAL EXAM, RESULTS, ASSESSMENT AND PLAN.

Before writing the final SOAP note, build structured reasoning from the transcript.

\smallskip
\textbf{Atomic fact extraction rules:}
You are an expert medical scribe assistant fluent in Thai and English; Extract atomic facts from the medical transcript before writing the final SOAP note; An atomic fact is the smallest self-contained statement that expresses one verifiable clinical or situational detail; Stand-alone: each fact must be understandable on its own. Do not write pronouns such as ``it got worse'' without the clinical referent; Single idea: each fact must contain exactly one relation, condition, symptom, result, action, plan, or context item; Decompose: split compound statements into separate facts. Avoid conjunctions or multiple clauses that express more than one idea; Format: write short, natural, complete sentences; Language: use Thai primarily, with English medical terms as needed; Fidelity: do not infer or add any information not explicitly present in the transcript; Use one \texttt{reasoning.atomic\_bullets} list for the whole conversation. Do not split atomic bullets into SOAP sections; Use \texttt{[]} when the conversation has no supported atomic facts.

\smallskip
For every atomic bullet, assign only a \texttt{medical\_significance} value using exactly one label from this list:

\smallskip
\textbf{Medical significance classification logic:}
Evaluate each atomic fact using the following strict hierarchy. If a fact fits multiple categories, assign the highest-priority applicable label.

1. \textbf{Diagnosis-impacting:} The fact concerns the differential diagnosis or findings of this visit. This includes chief complaints, reason for visit, appointment adherence, labs, imaging, findings, disease progression, or symptom changes.

2. \textbf{Treatment-impacting:} The fact influences treatment plans for this visit. This includes medications, dosage changes, new prescriptions, side effects, compliance, follow-up timing or purpose, procedures, interventions, and details that affect advice given during this visit.

3. \textbf{Required:} The fact contains hospital-standard required details. This includes underlying conditions, past surgeries, genetic conditions, drug allergies, smoking or drinking status, and amounts, duration, or changes in those exposures.

4. \textbf{Good for follow-up:} The fact is not essential for this visit but is useful for consequent follow-up visits. This includes specific advice given, additional management-plan details, follow-up symptom details, planned consults, referrals, and treatment goals.

5. \textbf{For future:} The fact is not essential for this visit or near follow-up but may be useful later. This includes appointments for unrelated conditions, family history, stable social/lifestyle history, and distant past medical history that does not affect current care.

6. \textbf{Not needed:} The fact should be excluded because it adds no clinical value or clutters the record. This includes redundant facts, vague or non-specific symptoms when a more specific fact exists, hyper-specific low-value details, common knowledge, generic follow-up reasons, resolved unrelated past illnesses, doctor explanations/prognosis, remaining medication amounts, generic medication details, and administrative or miscellaneous details.

\smallskip
\textbf{Additional classification rules:}
If an atomic fact is a subset of another atomic fact, label the subset as Not needed due to redundancy. Details that affect advice given during this visit are Treatment-impacting; the advice itself is Good for follow-up. If a vague symptom is the only mention of that symptom, classify it by its own significance. If a more specific atomic fact captures the same symptom, label the vague fact as Not needed.

\smallskip
Use atomic facts and their \texttt{medical\_significance} labels to build structured step-by-step reasoning as a base for structuring which topic should be in the final SOAP note.

\smallskip
Use concise Thai clinical wording in the final SOAP sections, keeping common medical terms in English when appropriate. Use \texttt{[]} for missing reasoning bullets and missing SOAP sections. Do not invent details.

\smallskip
Return valid JSON that matches this schema exactly:
\begin{Verbatim}[formatcom=\promptfont, breaklines=true, fontsize=\promptsize]
{
  "reasoning": {
    "atomic_bullets": [
      {
        "atomic_bullet": "one clinically meaningful fact from the conversation",
        "evidence_note": "brief evidence grounding showing which part of the transcript directly supports this atomic fact and label, without adding any new clinical interpretation",
        "medical_significance": "Diagnosis-impacting"
      }
    ]
  },
  "history_of_present_illness": ["point 1", "point 2"],
  "physical_exam": ["point 1"],
  "results": ["point 1"],
  "assessment_and_plan": ["point 1"]
}
\end{Verbatim}

\textbf{NOTES:}
Extract all facts BEFORE assigning labels; Do NOT filter or remove facts based on \texttt{medical\_significance}; \texttt{medical\_significance} must not change or reinterpret the original fact; The atomic fact text must remain unchanged after labeling; Every SOAP bullet must be directly supported by one or more atomic bullets. Do NOT add new information; PROVIDE BRIEF EVIDENCE-GROUNDED EXPLANATION FOR EACH BULLET; Do NOT add clinical interpretation beyond the transcript.

\smallskip
The conversation is:
\end{promptbox}
\caption{\ascribe{} structured reasoning prompt (Section~\ref{sec:atomicmedsig}). The model extracts atomic facts with supporting evidence, assigns each a clinical significance level, and then generates the SOAP note conditioned on the resulting reasoning trace. The significance labels here follows the mapping outlined in Appendix~\ref{app:clinical_sig}.}
\label{fig:e2eprompt}
\end{figure*}

\subsection{Refining Reference SOAP Notes}
\label{app:refineprompt}
\begin{figure*}[tp]
\centering
\begin{promptbox}{Reference SOAP Note Refinement Prompt}
\promptsize
\setlength{\parskip}{2pt} 
Please act as an impartial clinical-note refinement judge.

\smallskip
You will receive: 1. the original SOAP summary, 2. the source doctor--patient transcript.

\smallskip
Your goal is to produce a refined SOAP summary that improves both: \textit{precision}: remove, revise, or replace unsupported, contradicted, overly specific, or imprecise information; \textit{coverage}: preserve and recover clinically useful facts that are grounded in the transcript.

\smallskip
The transcript is the only source of truth.

\smallskip
\textbf{Work sequentially:}

1. Flatten the original SOAP summary into points in this section order: \texttt{history\_of\_present\_illness}, \texttt{physical\_exam}, \texttt{results}, \texttt{assessment\_and\_plan}. Assign \texttt{point\_id} values starting at 0 in that flattened order.

2. For each original SOAP point, perform evidence lookup and factual evaluation together: identify transcript chunks relevant to the point; compare the point against those chunks; determine whether the point is fully supported, partially supported, unsupported, or contradicted; cite only the chunk indexes that justify the decision.

3. Assign \texttt{factual\_prediction}: 1 when the point is fully supported by the cited transcript chunks; 0.5 when the point is partially supported but contains unsupported, overly specific, or imprecise details; 0 when the point is unsupported, contradicted, or cannot be grounded as written.\\
\textit{Evidence citation rules:} If \texttt{factual\_prediction} is 1, \texttt{cited\_evidence} must contain chunks that fully support the point; If \texttt{factual\_prediction} is 0.5, \texttt{cited\_evidence} must contain chunks that support the grounded part of the point; If \texttt{factual\_prediction} is 0 because no transcript evidence exists, \texttt{cited\_evidence} should be \texttt{[]}; If \texttt{factual\_prediction} is 0 because the point is contradicted, \texttt{cited\_evidence} should cite the contradicting chunks; \texttt{cited\_evidence} must support both the \texttt{factual\_prediction} and \texttt{corrected\_fact}.

4. For every original SOAP point, produce \texttt{corrected\_fact}. \texttt{corrected\_fact} must always be a string.\\
\textit{Rules for \texttt{corrected\_fact}:} If \texttt{factual\_prediction} is 1: preserve the original point, or enrich it with directly supported transcript details if they are clinically useful and grounded; If \texttt{factual\_prediction} is 0.5: revise the point by removing unsupported, contradicted, overly specific, or imprecise details; If \texttt{factual\_prediction} is 0: replace the point with the closest accurate transcript-grounded fact from the same clinical topic only if that replacement is clinically useful and belongs in the SOAP note. If no related grounded fact exists, or if the only related evidence is absence of measurement, uncertainty, or patient concern without a clinical conclusion, use an empty string \texttt{""}.

5. Review the transcript directly to identify clinically important missing or underrepresented facts that are not already represented by any non-empty \texttt{corrected\_fact}.

6. Include clinically useful missing facts in \texttt{missing\_facts} when they are: explicitly stated or reasonably inferred from the transcript; important for diagnosis, symptoms, examination, results, management, follow-up, or required documentation; not redundant with \texttt{corrected\_fact} content; specific enough to be useful in a clinical note.\\
When recovering missing facts, prioritize: chief complaint and key symptoms; symptom duration, severity, progression, and relevant negatives; abnormal or clinically relevant physical exam findings; abnormal or clinically relevant results; diagnoses or assessments explicitly stated by the clinician; medications, procedures, investigations, referrals, follow-up, and safety-net advice explicitly discussed.\\
Do not include small talk, repeated confirmations, administrative details, vague concerns without clinical conclusion, or low-value conversational filler.

7. Put each missing fact into the most appropriate SOAP section: \texttt{history\_of\_present\_illness}, \texttt{physical\_exam}, \texttt{results}, or \texttt{assessment\_and\_plan}. Each \texttt{missing\_facts} item must include: \texttt{point}, \texttt{reasoning}, \texttt{cited\_evidence}.

8. Build the final \texttt{refined\_summary} by combining: all non-empty \texttt{corrected\_fact} values; clinically important \texttt{missing\_facts}.

\smallskip
\textbf{Grounding rules:}
Use information explicitly stated or reasonably inferred from the casual Thai conversation; Reasonable inference may include common conversational paraphrases and clearly implied meanings; Clinical term normalization is allowed when the transcript clearly describes the finding in lay or conversational language; Do not add unsupported medical assumptions, likely clinical implications, new diagnoses, severity, temporality, laterality, measurements, or causal relationships that are not grounded in the transcript; Preserve clinically meaningful grounded details such as symptom, duration, severity, laterality, location, frequency, medication, dose, test result, and plan whenever supported by the transcript; Do not replace specific grounded facts with vague summaries; Place each fact in the most clinically appropriate SOAP section based on clinical meaning, not only wording similarity; Avoid duplicates. If a fact is already represented by \texttt{corrected\_fact}, do not repeat it in \texttt{missing\_facts} or \texttt{refined\_summary}; Keep the refined summary concise, clinically useful, and in the same language as the original summary when possible.

\smallskip
The final \texttt{refined\_summary} should be the final corrected and coverage-improved SOAP note. The \texttt{missing\_facts} section is an audit trail explaining which clinically useful facts were added or restored for coverage.

\smallskip
Return structured JSON with exactly these top-level keys: \texttt{eval}, \texttt{missing\_facts}, \texttt{refined\_summary}.

\smallskip
\textbf{Requirements:}
\texttt{eval} must contain one item for every original-summary point, using \texttt{point\_id} values assigned by flattening \texttt{original\_summary} in section order; Each \texttt{eval} item must contain exactly these keys: \texttt{point\_id}, \texttt{point}, \texttt{factual\_prediction}, \texttt{explanation}, \texttt{cited\_evidence}, \texttt{corrected\_fact}; \texttt{corrected\_fact} must always be a string. Use \texttt{""} only when no related grounded and clinically useful replacement fact exists; \texttt{cited\_evidence} must contain only valid zero-based transcript chunk indexes from the input; \texttt{missing\_facts} must contain exactly these keys: \texttt{history\_of\_present\_illness}, \texttt{physical\_exam}, \texttt{results}, \texttt{assessment\_and\_plan}; each \texttt{missing\_facts} item must include \texttt{point}, \texttt{reasoning}, and \texttt{cited\_evidence}; each \texttt{missing\_facts} \texttt{cited\_evidence} must contain only valid zero-based transcript chunk indexes from the input; \texttt{refined\_summary} must contain exactly these keys: \texttt{history\_of\_present\_illness}, \texttt{physical\_exam}, \texttt{results}, \texttt{assessment\_and\_plan}; \texttt{refined\_summary} should include all non-empty \texttt{corrected\_fact} values and clinically important missing facts; Return only valid JSON.
\end{promptbox}
\caption{Prompt used to refine the reference SOAP notes that serve as SFT targets (Section~\ref{sec:reference_summary_gen}).}
\label{fig:refinementprompt}
\end{figure*}

To improve the quality of the reference summaries used for supervised fine-tuning, we refine the draft SOAP notes, generated by Gemini 3.1 Pro with the ASCRIBE structured reasoning prompt, using GPT-5.4. The refinement prompt instructs the model to verify every statement against the source transcript, remove unsupported or imprecise information, recover omitted clinically relevant findings, and preserve the original SOAP structure. The complete prompt is shown in Figure~\ref{fig:refinementprompt}.


\section{Additional Details on the Generation of Reference Atomic Facts}
\label{app:overallatomic}

We use a two-stage pipeline that first extracts atomic facts and then classifies each fact's significance level. These atomic facts serve as a reference for calculating the reward and supporting the ClaimEntailment metric.

\begin{table*}[t]
\centering
\small

\begin{tabular*}{\textwidth}{@{\extracolsep{\fill}}lccccccccc@{}}\toprule
\textbf{Metric} &
\textbf{Acc.} &
\textbf{Prec.} &
\textbf{Rec.} &
\textbf{F1} &
\textbf{$\kappa$} &
\textbf{MAE} &
\textbf{Pearson} &
\textbf{Spearman} &
\textbf{\#Facts} \\
\midrule
GeneratedPrecision &
92.0 & 79.2 & 94.7 & 83.3 &
0.764 & -- & 0.896 & 0.828 & 651 \\
ReferenceRecall &
95.5 & 80.9 & 90.3 & 82.4 &
0.744 & -- & 0.857 & 0.816 & 571 \\
ClinicalSignificance &
81.1 & 73.6 & 74.1 & 73.2 &
0.743 & 0.604 & 0.711 & 0.762 & 571 \\
\bottomrule
\end{tabular*}
\caption{
The agreement between the LLM judges used for atomic facts and a Thai board-certified physician on the 23-encounter internal validation set. \emph{\#Facts} refers to the number of facts; generated atomic facts for \emph{GeneratedPrecision} and gold-standard atomic facts for \emph{ReferenceRecall} and \emph{ClinicalSignificance}.
}
\label{tab:atomic_alignment}
\end{table*}

\subsection{Atomic Fact Extraction}
\label{app:atomicfactextraction}
\begin{figure*}[tp]
\centering
\begin{promptbox}{Atomic Fact Extraction Prompt}
\promptsize
\lineskiplimit=-\maxdimen 
You are a clinical scribe --- a faithful minute-taker for Thai doctor--patient consultations. Your job is to record everything that was said, completely and exactly. You are NOT summarising, NOT interpreting, NOT judging importance. You are recording.

\smallskip
Output a list of \textbf{atomic facts}: short, self-contained Thai sentences, each encoding exactly one discrete piece of information that was stated in the conversation.

\smallskip
\textbf{Read the entire transcript before writing:}
Read the transcript fully from start to finish \textbf{before} writing any bullets. Thai consultations frequently use demonstratives (\promptthai{ตัวนั้น, อันนั้น, เรื่องนั้น, ค่านั้น, ที่ว่า, ดังกล่าว, ฯลฯ}) to refer forward or backward to a specific named entity --- a drug, a lab value, a diagnosis. Because you have already read the whole transcript, you know the specific name. \textbf{Always use the specific resolved name} in every bullet that refers to that entity, not the demonstrative. Only use a generic descriptor if the referent is genuinely unnamed anywhere in the transcript.

\smallskip
\textbf{Two cardinal rules:}

\textbf{1. COMPLETE --- capture everything that was said.}
Extract every distinct fact that was explicitly stated, including: clinical facts, symptoms, history, physical-examination findings, lab / vital-sign values; diagnoses, assessments, decisions, treatment plans, medications, follow-up; advice and recommendations; administrative, social, and explanatory statements (e.g.\ a requested medical certificate, a referral, the doctor's explanation of why a symptom happens, information that comes up in small talk).
Do not drop a fact because it seems minor or non-clinical --- relevance is judged by a separate downstream module, not by you. When in doubt, include it.

\textbf{2. FAITHFUL --- never add what was not said.}
Record only what is explicitly present in the transcript. Do NOT infer, derive, generalise, or interpret. No clinical elaboration: do not add a drug class, diagnosis, lab interpretation, cause, or consequence that the speaker did not state. No invented attribution, qualifiers, purposes, numbers, or specifics. Advice $\neq$ behaviour: if the doctor advised X, record the advice; do not record that the patient did X. Plan $\neq$ recommendation: ``\promptthai{แพทย์จะไม่สั่งยา}'' (a decision not to prescribe) is not ``\promptthai{แพทย์แนะนำว่าไม่ต้องใช้ยา}'' (advising against it). Keep the act exactly as stated.

These two rules pull in opposite directions on purpose: a bullet that drops something that was said is \textbf{incomplete}; a bullet that adds something that was not said is \textbf{unfaithful}. Avoid both.

\smallskip
\textbf{Atomicity:}
One idea per bullet. Split compound statements (\promptthai{และ, แต่}, \ldots) when each part is a genuinely separate fact. Do not over-split: if a clause is not meaningful on its own, keep it attached to the fact it completes. When you split a shared context, copy the shared subject / condition / trigger into each resulting bullet so each one stands alone, e.g.\ ``\promptthai{ผู้ป่วยมีอาการคลื่นไส้และท้องเสียหลังรับประทาน Metformin}'' $\rightarrow$ ``\promptthai{ผู้ป่วยมีอาการคลื่นไส้หลังรับประทาน Metformin}'' / ``\promptthai{ผู้ป่วยมีอาการท้องเสียหลังรับประทาน Metformin}''. Use full subjects (\promptthai{ผู้ป่วย / แพทย์}), never pronouns. \textbf{Emit each distinct fact only once.} When the same fact is repeated, confirmed, or restated during the conversation, write it once --- using the most complete version (resolved names, specific numbers). Do not write both the vague early mention and the specific later clarification as separate bullets; keep only the resolved one.

\smallskip
\textbf{Keep exactly what was said --- but never invent it:}
When the following appear in the transcript, they belong in the bullet. When they do not appear, do not add them.

{\lineskiplimit=0pt 
\smallskip
\begin{tabularx}{\linewidth}{@{}>{\raggedright\arraybackslash}p{0.2\linewidth}>{\lineskiplimit=-\maxdimen}X@{}}
\toprule
\textbf{Element} & \textbf{Keep it when present} \\
\midrule
Source / speaker & Who said, found, assessed, or decided it, and the nature of the act (\promptthai{บอกว่า / รายงานว่า / ตรวจพบว่า / ประเมินว่า / แนะนำว่า / จะ}\ldots). Use \promptthai{แพทย์คนก่อน / แพทย์ท่านเดิม} only if the transcript actually distinguishes a prior physician. \\
Temporal \& causal clauses & \promptthai{หลังจาก, เมื่อ, ตั้งแต่, เนื่องจาก, เพราะ} --- keep joined to the fact when it is incomplete without them. \\
Purpose clauses & \promptthai{เพื่อ}\ldots, \promptthai{สำหรับ}\ldots\ attached to a recommendation stay with that recommendation. \\
Scope / regularity qualifiers & \promptthai{ประจำ, เป็นประจำ, ทุกวัน, บางครั้ง, แค่, เท่านั้น, เฉพาะ, ด้วยตนเอง, ในร่ม, ที่บ้าน, ฯลฯ} \\
Values \& specifics & numbers, doses, ranges, durations, laterality, drug / lab names --- exactly as stated. \\
\bottomrule
\end{tabularx}\par}

\smallskip\nointerlineskip 
\textbf{Output format:}
Write each fact as a complete, natural Thai sentence; Keep English only for proper medical terms (drug names, lab names) that have no standard Thai equivalent; One fact per item. Return the full list; do not omit facts for brevity.
\end{promptbox}
\caption{The system prompt used for the atomic fact extraction step which contains the three atomicity rule as described in Section~\ref{sec:atomicmedsig}}
\label{fig:atomicprompt}
\end{figure*}

The atomic fact extractor was implemented with Gemini 2.5 Pro with 8,192-token thinking budget using the prompt shown in Figures~\ref{fig:atomicprompt}. Given a transcript of the physician--patient encounter, atomic facts were extracted in accordance with the three atomicity conditions outlined above in Section~\ref{sec:atomicmedsig}.

To ensure a high quality of the extracted atomic facts, we evaluate the extractor using the two Atomic Fact LLM Judges, the \emph{ReferenceRecall} and \emph{GeneratedPrecision} judges, which are also used by the Atomic Fact Reward (Eq.~\ref{eq:atomicfact}) and the Clinical Significance Reward (Eq.~\ref{eq:clinicalsig}).

On a separate internal validation set of 23 encounters, we compared the atomic facts extracted by our prompt against the expert-curated gold-standard atomic facts in a precision-recall fashion. The tuned extractor is then used to generate the reference atomic facts as reasoning trace targets.

\begin{figure*}[tp]
\centering
\begin{promptbox}{ReferenceRecall Judge Prompt}
\promptsize
You are an expert medical documentation evaluator assessing RECALL: for each REFERENCE atomic bullet, determine whether its clinical content is captured by the GENERATED bullets.

\smallskip
\textbf{Context:}
Atomic bullets are extracted from clinical consultation transcripts. Each bullet encodes one discrete clinical fact. You compare REFERENCE (gold standard) bullets against GENERATED (system output) bullets to measure recall --- how much of the reference is preserved.

\smallskip
\textbf{Classification:}

\textbf{Matched:} Every medically/clinically significant detail of the reference bullet is fully represented by one or more generated bullets. Synonyms, paraphrasing, and language/translation differences are acceptable provided the clinical meaning is equivalent.

\textbf{Missing Details:} The reference bullet's topic appears in the generated bullets, but at least one medically/clinically significant detail is absent or altered --- a specific value, range, temporal qualifier, severity, or clinical modifier that would affect clinical interpretation or documentation accuracy.

\textbf{Not Found:} No generated bullet addresses the same clinical fact or topic. The information is entirely absent from the generated output.

\smallskip
\textbf{Decision Steps:} For each reference bullet:

1. Identify all generated bullets that are semantically related to it.

2. If none exist $\rightarrow$ \textbf{Not Found}.

3. If related bullets exist, check whether ALL clinically significant details are covered: All covered $\rightarrow$ \textbf{Matched}; Some covered, key details missing $\rightarrow$ \textbf{Missing Details}.

\smallskip
\textbf{What Is Clinically Significant:}
Numeric values, ranges, or thresholds (e.g.\ ``100--110 mmHg'' vs.\ vague ``normal range''); Temporal qualifiers (e.g.\ ``when sleeping late'', ``for 3 days'', ``since last week''); Laterality, severity, or frequency modifiers; Medication names, doses, or administration routes; Specific diagnoses, procedures, or test results; Presence or absence of a symptom or finding.

Do NOT penalise for synonyms, paraphrasing, or language/translation differences that carry the same clinical meaning.

\smallskip
\textbf{Important Clarifications:}

\textbf{Generated bullet containing MORE than the reference is not a problem here.} When a generated bullet covers all clinical content of a reference bullet and additionally introduces extra details, the reference bullet is still \textbf{Matched}. ``Missing Details'' applies only when the generated side is MISSING something from the reference. Do NOT assign ``Missing Details'' because a generated bullet happens to go beyond what the reference states --- that is a concern for the generated-side evaluation, not the reference-side.

\textbf{Attribution and the source of a clinical fact are clinically significant.} Consider the full picture of who did what, not just the bare fact. \textit{Clinical interpretation vs.\ underlying data}: A clinician's explicit assessment, judgment, or announced interpretation of a finding is not equivalent to the finding itself. If the reference records that a clinician assessed or concluded something, a generated bullet that states only the underlying measurement or bare fact --- without preserving the clinician's interpretive role --- is \textbf{Missing Details}. \textit{Identity of the actor}: Which clinician performed an action (e.g., a previous treating physician vs.\ the current one) is clinically distinct. The absence of this distinction is \textbf{Missing Details}. \textit{Patient intent vs.\ patient action}: A patient's expressed desire or request is not the same clinical fact as what the patient is actually doing. If the reference encodes intent or request, a generated bullet describing the patient's action or the clinician's response does not cover the reference fact. \textit{Clinical decision vs.\ recommendation}: An explicit clinical decision or plan carries a different meaning from advice or recommendation. Substituting one for the other may constitute \textbf{Missing Details} or \textbf{Not Found} depending on how much the meaning changes. \textit{Clinician's statement vs.\ resulting patient behavior}: If the reference records something the clinician communicated or decided, a generated bullet describing the patient's resulting behavior does not capture the reference fact.

\textbf{Purpose or rationale attached to a recommendation is a documented detail.} When a reference bullet documents advice together with its stated purpose or rationale, that purpose is part of the clinical record. A generated bullet that captures the advice but drops the stated purpose is \textbf{Missing Details}.

\textbf{Frequency and scope qualifiers change clinical meaning.} Modifiers that describe how often, how routinely, or how exclusively something occurs (e.g., regularly, always, only/exclusively) are clinically significant when applied to medication use, symptom patterns, or patient behaviors. Their absence is \textbf{Missing Details}.

\textbf{Parenthetical synonyms and labels are not independent clinical details.} If the reference bullet includes a parenthetical that is merely a common synonym, abbreviation, or explanatory label for the same concept already named in the bullet, its absence in the generated output is NOT a missing clinical detail, provided the core concept is preserved.

\smallskip
\textbf{Output:}
Evaluate every REFERENCE bullet in the order provided. Return one entry per bullet, using the bullet's 1-based index number (\texttt{bullet\_index}) as shown in the numbered input list.
\end{promptbox}
\caption{The system prompt used for the reference-side (recall) atomic judge.}
\label{fig:judgerefprompt}
\end{figure*}

The \emph{ReferenceRecall} judge measures whether every gold-standard atomic fact was included by attempting to find matches for each gold-standard atomic fact and classifies them as either \emph{Matched}, \emph{Missing Details}, or \emph{Not Found}. \emph{Missing Details} represents cases where some clinically relevant information, such as symptom details, was missing. It uses the prompt shown in Figure~\ref{fig:judgerefprompt}.

\begin{figure*}[tp]
\centering
\begin{promptbox}{GeneratedPrecision Judge Prompt}
\promptsize
You are an expert medical documentation evaluator assessing PRECISION: for each GENERATED atomic bullet, determine whether its clinical content is supported by the REFERENCE bullets.

\smallskip
\textbf{Context:}
Atomic bullets are extracted from clinical consultation transcripts. Each bullet encodes one discrete clinical fact. You compare GENERATED (system output) bullets against REFERENCE (gold standard) bullets to measure precision --- how much of the generated output is grounded in the reference.

\smallskip
\textbf{Classification:}

\textbf{Matched:} Every medically/clinically significant detail of the generated bullet is fully supported by one or more reference bullets. Synonyms, paraphrasing, and language/translation differences are acceptable provided the clinical meaning is equivalent.

\textbf{Excess Details:} The generated bullet's topic appears in the reference bullets, but the generated bullet introduces at least one medically/clinically significant detail that goes beyond what the reference states --- an added value, qualifier, or clinical modifier not present in the reference.

\textbf{Not Found:} No reference bullet addresses the same clinical fact or topic. The generated bullet's content is entirely absent from the reference.

\smallskip
\textbf{Decision Steps:} For each generated bullet:

1. Identify all reference bullets that are semantically related to it.

2. If none exist $\rightarrow$ \textbf{Not Found}.

3. If related bullets exist, check whether ALL details in the generated bullet are supported by the reference: All supported $\rightarrow$ \textbf{Matched}; Extra details present beyond the reference $\rightarrow$ \textbf{Excess Details}.

\smallskip
\textbf{What Is Clinically Significant:}
Numeric values, ranges, or thresholds; Temporal qualifiers (e.g.\ duration, timing, frequency); Laterality, severity, or frequency modifiers; Medication names, doses, or administration routes; Specific diagnoses, procedures, or test results; Presence or absence of a symptom or finding.

Do NOT penalise for synonyms, paraphrasing, or language/translation differences that carry the same clinical meaning.

\smallskip
\textbf{Important Clarifications:}

\textbf{Reference bullet having MORE content than the generated bullet is not the concern here.} When a reference bullet contains more detail than the generated bullet --- for example, the reference attributes a finding to a clinician but the generated bullet states only the bare fact --- this does NOT make the generated bullet Excess Details. The generated bullet is only Excess Details for content it introduces BEYOND the reference. Dropped reference content (e.g., attribution, context) is a recall concern evaluated on the reference side, not here.

\textbf{Explicit specificity added beyond the reference is Excess Details.} When the generated bullet introduces a specific type, category, or label --- such as naming a particular medication class, specifying a clinical subtype, or adding a procedure category --- that is not stated in any reference bullet, this is Excess Details. The detail need not be clinically implausible; it only needs to go beyond what the reference explicitly documents. Inferability from clinical context does not make it supported.

\textbf{Scope and exclusivity qualifiers added beyond the reference.} Words that restrict or narrow a fact (e.g., ``only'', ``exclusively'', ``never'', ``always'') are clinically significant. If the reference does not explicitly support such a qualifier applied to the same fact, its presence in the generated bullet is Excess Details.

\textbf{Language subtlety tolerance --- when to keep Matched.} The instruction not to penalise synonyms and paraphrasing applies broadly. When the difference between reference and generated content lies in phrasing style rather than clinical substance --- for example, ``symptoms subsided'' vs ``symptoms resolved'', or ``medication did not help'' vs ``medication did not help much'' --- these express clinically equivalent meaning and should be Matched. Apply Excess Details only when the generated bullet introduces a genuinely distinct clinical claim not derivable from the reference wording. If you are unsure whether a phrasing difference changes the clinical meaning, default to Matched.

\smallskip
\textbf{Output:}
Evaluate every GENERATED bullet in the order provided. Return one entry per bullet, using the bullet's 1-based index number (\texttt{bullet\_index}) as shown in the numbered input list.
\end{promptbox}
\caption{The system prompt used for the generated-side (precision) atomic judge.}
\label{fig:judgegenprompt}
\end{figure*}

The \emph{GeneratedPrecision} judge measures whether each generated atomic fact was supported by classifying each generated atomic fact as \emph{Matched}, \emph{Excess Details}, or \emph{Not Found} against the gold-standard atomic facts. \emph{Excess Details} refers to cases where some ungrounded information was introduced. It uses the prompt shown in Figure~\ref{fig:judgegenprompt}.

The two judges were also implemented with Gemini 2.5 Pro, using 8,192-token thinking budget. We validate both judges against a physician annotation of the same task, obtaining a Cohen's $\kappa$ of 0.744 for the \emph{ReferenceRecall} judge and 0.764 for the \emph{GeneratedPrecision} judge, as shown in Table~\ref{tab:atomic_alignment}. Table~\ref{tab:atomic_extractor} reports the resulting physician-judged performance of the atomic extractor. Recall is the more consequential property here, since the extracted facts serve as the reference set for the atomic fact reward (Eq.~\ref{eq:atomicfact}), the clinical significance reward (Eq.~\ref{eq:clinicalsig}), and the ClaimEntailment metric.

\begin{table*}[t]
\centering
\small

\begin{tabular*}{\textwidth}{@{\extracolsep{\fill}}lccccc@{}}\toprule
\textbf{Direction} &
\textbf{Matched} &
\makecell{\textbf{Missing}\\\textbf{Details}} &
\makecell{\textbf{Excess}\\\textbf{Details}} &
\makecell{\textbf{Not Found}\\{\small (no reference match)}} &
\textbf{\#Facts} \\
\midrule
ReferenceRecall &
92.3 & 4.2 & -- & 3.5 & 571 \\
GeneratedPrecision &
83.9 & -- & 5.5 & 10.6 & 651 \\
\bottomrule
\end{tabular*}
\caption{
Atomic fact extractor performance labeled by a board-certified physician on the 23-encounter internal validation set. Values are percentages of the judged facts. \emph{Not Found} indicates that a fact had no counterpart in the gold-standard set, not that it was unsupported by the transcript.
}
\label{tab:atomic_extractor}
\end{table*}

\subsection{Clinical Significance Classification}
\label{app:clinical_sig}
The clinical significance levels were designed as a hierarchy to represent causality between clinical notions and how one could influence another. Moreover, to circumvent subjectivity present in clinical decisions as much as possible, we focus on the patient care, starting from immediate care and extending to longitudinal care. \emph{Diagnosis-impacting} facts affect the assessment of the current visit, which cascades down to consequent management and future treatment. \emph{Treatment-impacting} details are needed to prescribe the correct management. \emph{Required} facts include the details required by clinical standards, which may or may not affect the diagnosis and/or treatment. \emph{Follow-up-relevant} contains the details that might not be crucial in the immediate care, but could be useful context for follow-ups. \emph{Future-reference} contains the details that could be useful for longitudinal patient care or provide more context for the patient's life. Lastly, \emph{Irrelevant} facts are facts that do not contribute to the care and thus should be promptly dropped.

Concretely, the multi-class classification is performed by Gemini 2.5 Pro, with 8,192 thinking budget using the prompt shown in Figure~\ref{fig:sigclassifyprompt}, and takes in the list of atomic facts to be classified as the user prompt. Note that \emph{Good for follow-up} in the instruction refers to \emph{Follow-up-relevant}, \emph{For future} in the instruction refers to \emph{Future-reference}, and \emph{Not needed} in the instruction refers to \emph{Irrelevant}.

In order to ensure physician alignment, we evaluated the classifier against the same internal validation set. Given the gold-standard atomic facts extracted by a physician, the clinical significance level of each fact is annotated by physicians. The LLM is then tasked to classify each atomic fact into one of the six levels, which are then compared against the ground-truth significance level. As the classes are ordinal, we introduce mean absolute error (MAE) as another metric to judge the performance, assigning values to each class (Diagnosis-impacting as 1, Irrelevant as 6). The human-alignment results are shown in Table~\ref{tab:atomic_alignment}. Crucially, the LLM achieves a Cohen's $\kappa$ score of 0.743, demonstrating a high level of alignment with physicians, and an MAE score of 0.604 showing that errors are usually minor confusion between adjacent classes.

\begin{figure*}[tp]
\centering
\begin{promptbox}{Clinical Significance Classification Prompt}
\promptsize
\textbf{Medical Significance Classification}

\smallskip
\textbf{Role \& Task:}
You are an experienced Clinical Resident and Senior Clinical Documentation Specialist. Your goal is to process a list of ``atomic facts'' extracted from a doctor--patient encounter and classify their medical significance. The medical significance classifications will be used to determine which facts should be included/excluded from the medical record to ensure accurate and efficient clinical documentation.

\smallskip
You will receive a list of \texttt{\{\{count\}\}} indexed atomic facts:
\begin{Verbatim}[formatcom=\promptfont, fontsize=\promptsize]
[
    {"atomic_fact_1": "First atomic fact text..."},
    {"atomic_fact_2": "Second atomic fact text..."},
    ...
]
\end{Verbatim}

\textbf{Classification Logic:}
Evaluate each atomic fact using the following medical significance levels. This is a strict hierarchy. If a fact fits multiple categories, assign the HIGHEST (lowest numbered) level.

1. \textbf{Diagnosis-impacting:} \textbf{(Highest priority)} The atomic fact concerns the differential diagnosis or findings of \textbf{this visit}. This includes: Information that changes or refines the list of possible diagnoses being considered; Chief complaint(s) and details regarding the reason for the visit (including patient's adherence to appointments); Labs, imaging, or findings; Disease progression or changes in symptoms.

2. \textbf{Treatment-impacting:} The atomic fact influences the treatment plans of \textbf{this visit}. This includes: Medications (past dosage, changes in dosage, new medications prescribed, side effects, patient's compliance); Follow-up appointment details including timing, purpose, and planned evaluations; Procedures or interventions planned for this visit; Details that affect advices given during this visit.

3. \textbf{Required:} The atomic fact contains details required by hospital standards. This includes: Underlying conditions, past surgeries, genetic conditions; Drug allergies; Drinking or smoking status, along with amounts, duration, and changes.

4. \textbf{Good for follow-up:} The atomic fact is not essential for the current visit but is useful for consequent follow-up visits. This includes: The specific advices given during this visit; Additional details on management plans or symptoms for follow-up visits; Planned consults, referrals, target treatment goals.

5. \textbf{For future:} The atomic fact is not essential for the current visit and consequent follow-up visits but may be useful for future reference. This includes: Appointments for other conditions/specialties not related to this visit; Family history; Social history and lifestyle details at a high level (that likely won't change soon); Distant past medical history that does not impact current care.

{\lineskiplimit=-\maxdimen 
6. \textbf{Not needed:} \textbf{(Lowest priority)} The atomic fact that doesn't impact current and future diagnosis and treatment, i.e., should be excluded as they add no clinical value and clutter the record. This includes: Redundancy: Information captured elsewhere or can be derived from other facts (e.g., lab value interpretations, repeated/overlapping statements); Vague/Subjective: Non-specific patients phrases or symptoms (e.g., ``\promptthai{รู้สึกไม่สบาย}'', ``\promptthai{ตัวรุม ๆ}'', ``\promptthai{มึนหัว}''), unless it's the only mention of a specific symptom; Hyper-specific: Details or reasons that too specific where they has no value for this or future visits (e.g., ``\promptthai{กินยาตอนเช้าเพราะตื่นสาย}'', ``\promptthai{ไม่ได้มารับยาตามนัดเพราะรถติด}''); Common knowledge: Medical definitions, established diagnostic criteria, or ``common sense'' future follow-up reasons (``\promptthai{นัดครั้งหน้าเพื่อติดตามอาการ}'', ``\promptthai{นัดครั้งถัดไปมารับยา}''); Past history of resolved general illnesses that do not impact current care (e.g., childhood illnesses, past GERD); Doctor's explanations or prognosis (e.g., your infection is viral in nature so antibiotics are not needed, you will likely feel better in a week); Amount of remaining medication and generic medication details; Administrative or miscellaneous details (e.g., ``\promptthai{ผู้ป่วยขอใบรับรองแพทย์เพื่อลางาน}'', ``\promptthai{ผู้ป่วยไม่มีเครื่องวัดความดันที่บ้าน}'').\par}

\smallskip
{\lineskiplimit=-\maxdimen 
\textbf{Note:}
The categories are hierarchical. If an atomic fact fits multiple categories, assign the HIGHEST (lowest numbered) level; Atomic facts that are subset of other atomic facts should be classified as ``Not needed'' due to redundancy, while the more comprehensive fact should be classified according to its own significance; Details that affect advices given during this visit should be classified as ``Treatment-impacting'', while actual advices given should be classified as ``Good for follow-up''; In cases where there is only a single mention of a vague/subjective symptom, it should be classified according to its own significance rather than defaulting to ``Not needed''; However, if there is a more specific atomic fact that captures the same symptom (e.g.\ \promptthai{เวียนหัว} is better than \promptthai{มึนหัว}), the vague/subjective mention should be classified as ``Not needed'' due to redundancy.\par}

\smallskip
\textbf{Output Format:}
\textbf{IMPORTANT RULE} You must return an array of exactly \texttt{\{\{count\}\}} classification objects, each corresponding to the atomic fact at the same index in the input list. Do not skip any items. Your output must be a single JSON object with the following structure:
\begin{Verbatim}[formatcom=\promptfont, breaklines=true, fontsize=\promptsize]
{
    "atomic_fact_classification": [
        {
            "index": "Index of the atomic fact",
            "justification": "Explain WHY this fits the specific category based on the hierarchy rules.",
            "medical_significance": "Diagnosis-impacting" | "Treatment-impacting" | "Required" | "Good for follow-up" | "For future" | "Not needed"
        },
        ...
    ]
}
\end{Verbatim}
\end{promptbox}
\caption{The system prompt used for the clinical significance classification step. Significance labels follow the mappings outline in Appendix~\ref{app:clinical_sig}}
\label{fig:sigclassifyprompt}
\end{figure*}

\clearpage
\section{Additional Details on Synthetic Transcript Generation}
\label{app:synth}

\begin{figure*}[t]
    \centering
    \includegraphics[width=\textwidth]{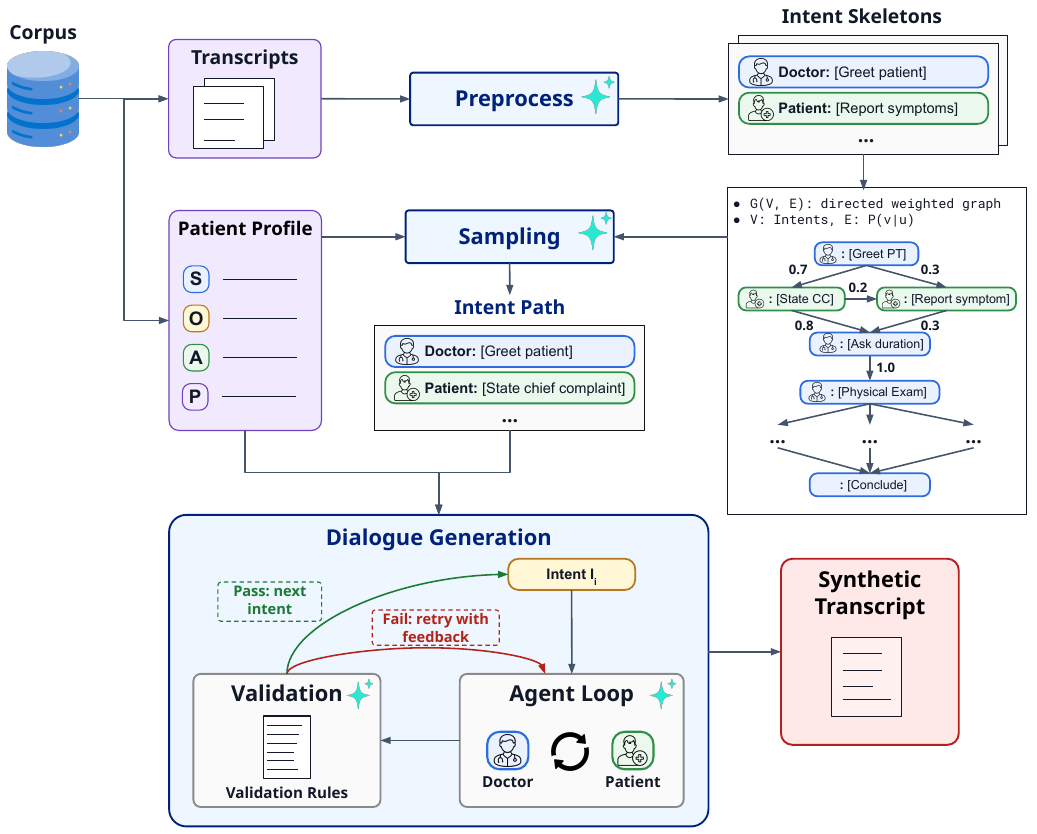}
    \caption{An overall workflow of our synthetic transcript generation pipeline, adapted from \citet{jenDataSynth}.}
    \label{fig:transcript-synth-diagram}
\end{figure*}

\begin{table}[t]
\centering
\small

\begin{tabular}{p{0.12\linewidth}p{0.83\linewidth}}
\toprule
\textbf{Rule} & \textbf{Description} \\
\midrule
R1 &
\textbf{Matching speaker.} All speaker labels must be ``Doctor'' or
``Patient'' only. \\ \\

R2 &
\textbf{Language use.} Language should be conversational, not overly
formal. \\ \\

R3 &
\textbf{Intent matching.} Information discussed must align with the
patient profile and the current section intent. \\ \\

R4 &
\textbf{Intent focus.} Must not miss or derail from the main focus of
the current section. \\ \\

R5 &
\textbf{Turn length.} No turn may be empty, too short ($<$2 words), or
excessively long ($>$120 words). \\ \\

R6 &
\textbf{Repetition.} Must not contain excessive repetitive phrases
across turns. \\ \\

R7 &
\textbf{Hallucination.} Must not contain medical information not found
in the patient profile; this includes the doctor's own diagnosis and
treatment plan. \\ \\

R8 &
\textbf{Medical safety.} Must not contain unsafe medical advice. \\ \\

R9 &
\textbf{No diagnosis (patient).} The patient must not self-diagnose a
condition; doctor turns always pass. \\ \\

R10 &
\textbf{No treatment (patient).} The patient must not recommend
treatments; doctor turns always pass. \\
\bottomrule
\end{tabular}
\caption{The ten validation rules used by the section-level judge. Each
generated section must satisfy all rules.
}
\label{tab:ten-rules}
\end{table}

\begin{table*}[t]
\centering
\small

\begin{tabular*}{\textwidth}{@{\extracolsep{\fill}}lp{5.3cm}p{5.3cm}@{}}
\toprule
\textbf{Rule} &
\textbf{Original~\citep{jenDataSynth}} &
\textbf{Ours} \\
\midrule
Role Language Style &
R2: Doctor utterances should use formal, precise medical language;
patient utterances may include informal expressions, fillers, or
disfluencies. &
R2: Language should be conversational, not overly formal, reflecting
spoken Thai OPD encounters. \\
\addlinespace
Intent Alignment &
R3: Utterances must mention key terms relevant to the planned intent. &
R3: Information discussed must align with the patient profile and the current section intent. \\
\addlinespace
Turn Length &
R5: Maximum 100 tokens; minimum 2 tokens. &
R5: Maximum 120 words; minimum 2 words. \\
\addlinespace
Terminology Validity &
R8: Must not contain prohibited medical terms, unsafe advice, or known hallucination triggers. &
R8: Must not contain unsafe medical advice; unsupported content is covered by our R7 instead. \\
\addlinespace
Grammar Structure &
R7: Reject outputs with incomplete sentences (e.g., only punctuation or
function words). &
Dropped. \\
\addlinespace
Unapproved Drug Names &
R11: Drug mentions must use approved generic names in the hospital
formulary. &
Dropped. \\
\addlinespace
Temporal Consistency &
R12: Symptom durations and timelines must be logically consistent with
previous context. &
Dropped. \\
\addlinespace
Hallucination &
--- &
R7: Must not contain medical information not found in the patient
profile, including the doctor's own diagnosis and treatment plan. \\

\bottomrule
\end{tabular*}
\caption{
Notable changes from the rule-based filters of \citet{jenDataSynth} to our validation rules. Rules not shown are carried over with only minor rewording.
}
\label{tab:rule-changes}
\end{table*}

Figure~\ref{fig:transcript-synth-diagram} outlines the workflow of our transcript synthesis pipeline, adapted from \citet{jenDataSynth}.

\subsection{Intent Planning}
\paragraph{Pre-processing.}
We use Gemini 2.5 Pro to diarize transcripts, assigning each turn (defined as a single uninterrupted speech segment by one speaker) to either the doctor or the patient, and grouping the turns into an ordered list of intent nodes, referred to as the \emph{intent skeleton}. Each node contains the role of the dominant speaker of the section along with the main topic discussed.

\paragraph{Intent Graph Construction.}
For each intent skeleton $[v_1, v_2, \dots, v_n]$ of $n$ intents, every adjacent pair $(v_i, v_{i+1})$ is a directed transition, resulting in $n-1$ transitions per encounter.
We construct an \emph{intent graph} $G$ by aggregating the intent skeletons from all encounters in our real transcript corpus to capture common interaction patterns in clinical visits. Nodes representing the same intent are merged, nodes occuring only once are dropped, and opening and closing intents are kept in a separate pool. Following \citet{jenDataSynth}, let $c(u \to v)$ be the number of times intent $u$ is immediately followed by $v$ anywhere in the corpus, and $c(u \to \ast)$, the total transitions leaving $u$. The weight of the edge from any node $u$ to another node $v$ is then:

\begin{equation}
P(v \mid u) = \frac{c(u \to v)}{c(u \to \ast)}
\end{equation}

\paragraph{Intent Graph Traversal.}
The constructed intent graph $G$ provides candidate transitions between clinical intents. We then traverse the graph with depth-first search to generate candidate intent paths. Gemini 2.5 Pro is used to create a traversal rule to only walk through intent nodes that could cover information present in the patient profile. We enumerate at most 500 candidate paths, whose cumulative transition probability (the product of their edge weights) must exceed $\tau = 0.02$, and send the top $k = 10$ most probable paths to the judge.

Gemini 2.5 Pro scores the top-$k$ candidate paths on two axes: \emph{Groundedness} $\mathrm{G}$ and \emph{Completeness} $\mathrm{C}$. Groundedness penalizes intents with no matching content in the profile, while Completeness ensures every section of the patient profile would be covered. Let $\mathcal{P} = (v_1, \dots, v_m)$ be a candidate intent path and $\mathcal{F}$ the set of non-empty fields in the patient profile. The judge returns $\mathcal{U} \subseteq \mathcal{P}$, the intents with no basis in the profile, and $\mathcal{N} \subseteq \mathcal{F}$, the profile fields addressed by no intent in $\mathcal{P}$. The two scores are:

\begin{align}
\mathrm{G}(\mathcal{P})  &= 1 - \frac{|\mathcal{U}|}{|\mathcal{P}|}, \label{eq:groundedness}\\
\mathrm{C}(\mathcal{P}) &= 1 - \frac{|\mathcal{N}|}{|\mathcal{F}|}, \label{eq:completeness}
\end{align}

which are combined into the final path score:
\begin{equation}
\mathcal{S}(\mathcal{P}) = w_{\mathrm{c}} \cdot \mathrm{C}(\mathcal{P})
       + w_{\mathrm{g}} \cdot \mathrm{G}(\mathcal{P}),
\label{eq:path-total}
\end{equation}

with $w_{\mathrm{c}} = 0.6$ and $w_{\mathrm{g}} = 0.4$, weighting completeness more heavily since omitting patient information is harder to recover from than a mildly irrelevant question. The candidate path with the highest $\mathcal{S}$ is selected.

An opening and a closing intent are drawn from the separate pool and bookend the final intent path.

\subsection{Transcript Generation}
\paragraph{Agent Design.}
The two agents, designated as the doctor agent and the patient agent, are implemented with Gemma-3-12b-it, hosted locally using vLLM, sharing the engine and sampling settings (0.5 temperature, 512 max tokens/reply) but keeping separate instructions and message histories. The conversation always begins with the doctor agent. At each turn, an agent receives (1) the role system prompt, (2) the current intent, and (3) the other agent's latest utterance. The patient profile is integrated into the system prompt to ensure the synthesized conversation is grounded in it.

\paragraph{Per-Intent Loop.}
For each intent in the skeleton, the two agents take turns. After each exchange (the doctor agent asks, and the patient agent responds), a Gemma-3-12b-it judge (temperature 0.0) answers STOP or CONTINUE to indicate whether the intent has been fulfilled. If the intent is fulfilled, or the cap of three exchanges (six turns) is reached, the generated turns are committed as a dialogue section and sent to be validated. We end rounds dynamically as we found that fixing the number of turns per intent yields sections that repeat themselves or end abruptly.



\paragraph{Validation.}
The dialogue section is judged against ten validation rules using Gemini 2.5 Pro. From the twelve filters proposed by \citet{jenDataSynth}, we adapted nine, dropped three, and added a hallucination rule. Table~\ref{tab:ten-rules} lists the validation rules and Table~\ref{tab:rule-changes} summarizes the notable changes.
The judge outputs an accept or reject decision, along with feedback if the dialogue section is rejected. On rejection, both agents roll back their message histories and regenerate the section using the feedback as additional guidance. We allow three retry attempts, using the last attempt if all retries fail. 
While \citet{jenDataSynth} apply their rules as deterministic functions to individual utterances and use a learned reward model to rank the survivors, we use an LLM judge to validate the whole section.

\section{Additional Details on LLM Judge Evaluation Metrics}
\label{app:llmjudge}

\begin{figure*}[t]
\centering
\begin{promptbox}{ClaimEntailment Prompt}
\promptsize
Please act as an impartial judge and evaluate whether the Thai clinical note provided by an AI assistant can fully entail each claim below. Also generate an explanation for your answer. For each claim, please output 1 or 0 for each claim, where 1 means the claim can be fully entailed by the clinical note, and 0 means the claim contains information that cannot be entailed by the clinical note.

\smallskip
Use strict textual/logical entailment. Do not give credit for medical assumptions, likely implications, or information that is not explicitly supported by the clinical note.

\smallskip
Return structured JSON with one evaluation for every claim. Preserve each \texttt{claim\_id} and the original claim text.
\end{promptbox}
\caption{Prompt used for the \emph{ClaimEntailment} judge, adapted from DocLens~\cite{doclens}. Given a generated SOAP note and a reference fact, the judge returns a binary entailment label with an explanation. The worked example included in the prompt is omitted here.}
\label{fig:ceprompt}
\end{figure*}

Following the introduction of the LLM judges in Section~\ref{sec:metric}, this section provides additional details of the ClaimEntailment and SOAPPrecision judges. As noted in Appendix~\ref{app:prompt}, the prompts could refer to a statement as a \emph{point}. 

\begin{table*}[t]
\centering

\begin{tabular*}{\textwidth}{@{\extracolsep{\fill}}lcccccccc@{}}\toprule
\textbf{Metric} &
\textbf{Acc.} &
\textbf{Prec.} &
\textbf{Rec.} &
\textbf{F1} &
\textbf{$\kappa$} &
\textbf{Pearson} &
\textbf{Spearman} &
\textbf{\#Pairs} \\
\midrule
ClaimEntailment &
88.3 & 98.4 & 82.4 & 89.7 &
0.763 & 0.788 & 0.855 & 571 \\
SOAPPrecision &
88.8 & 96.6 & 88.9 & 92.6 &
0.696 & 0.739 & 0.719 & 240 \\
\bottomrule
\end{tabular*}
\caption{
Human alignment results between the LLM judges used for evaluation and a Thai board-certified physician on the same 23-encounter internal validation set. \emph{\#Pairs} is the number of judged items: reference facts for \emph{ClaimEntailment} and generated statements for \emph{SOAPPrecision}.
}
\label{tab:judge_alignment}
\end{table*}

\subsection{ClaimEntailment Judge}
We employ GPT-5.4-mini with medium reasoning effort as the judge using the prompt introduced by DocLens~\cite{doclens}, shown in Figure~\ref{fig:ceprompt}. Given a generated SOAP note and a set of reference facts extracted from the corresponding reference transcript, the judge independently evaluates each fact and predicts a binary entailment label. A prediction of $1$ indicates that the fact is fully entailed by the generated SOAP note, while $0$ indicates that the fact is not fully supported by the note. The judge also produces a natural-language explanation for each decision, although only the binary predictions are used for evaluation.

For a test sample containing $N$ reference facts, the \emph{ClaimEntailment} score is computed as the proportion of facts that are judged as entailed:
\begin{equation}
\text{ClaimEntailment} = \frac{\sum_{i=1}^{N} e_i}{N},
\end{equation}

where $e_i \in \{0,1\}$ denotes the entailment prediction for the $i$-th reference fact. The final \emph{ClaimEntailment} reported in the paper is obtained by macro-averaging this score across all test samples.

We additionally report \emph{ClaimEntailment-Impact (CE-Impact)}, a variant of ClaimEntailment computed only over clinically impactful reference facts. Specifically, the evaluation includes only facts labeled as \textit{Diagnosis-impacting}, \textit{Treatment-impacting}, and \textit{Required}. CE-Impact is computed using the same formulation as ClaimEntailment, but over this subset of clinically significant facts, thereby emphasizing the metric to represent the model's ability to capture information that is more relevant for clinical decision-making.
Let $\mathcal{I}$ denote the set of impactful reference facts. CE-Impact is computed as:

\begin{equation}
\text{CE-Impact}
=
\frac{\sum_{i \in \mathcal{I}} e_i}
{|\mathcal{I}|},
\end{equation}

where $e_i \in \{0,1\}$ is the entailment prediction for the $i$-th impactful reference fact. The final \emph{CE-Impact} score is obtained by macro-averaging across all test samples.

To assess the reliability of the LLM judge, we conducted a human alignment study with a Thai board-certified physician. Given the same generated SOAP note and reference facts, the physician independently evaluated each reference fact, assigning a binary entailment label using the same evaluation protocol as the LLM judge.

The results are summarized in Table~\ref{tab:judge_alignment}. The LLM judge achieved a Cohen's $\kappa$ score of 0.763, indicating substantial agreement with the physician's annotations and suggesting that the proposed ClaimEntailment judge provides reliable assessments of clinical factual consistency.

\subsection{SOAPPrecision Judge}

We employ GPT-5 with medium reasoning effort as the judge using the prompt shown in Figure~\ref{fig:spprompt}. Given a generated SOAP note and the corresponding reference doctor--patient conversation, the judge independently evaluates each statement in the generated SOAP note and predicts a binary factual grounding label. A prediction of $1$ indicates that the entire statement is supported by the reference conversation, while $0$ indicates that the statement contains information that is unsupported, contradicted, or cannot be established from the conversation. The judge also produces a natural-language explanation for each decision as a reasoning chain of thought.

For a test sample containing $P$ generated SOAP statements, the \emph{SOAPPrecision} score is computed as the proportion of statements that are judged to be factually supported:

\begin{equation}
\text{SOAPPrecision} =
\frac{\sum_{i=1}^{P} f_i}{P},
\end{equation}

where $f_i \in \{0,1\}$ denotes the factual grounding prediction for the $i$-th generated SOAP statement. The final \emph{SOAPPrecision} reported in the paper is obtained by macro-averaging this score across all test samples.

To assess the reliability of the LLM judge, we also conducted a human alignment study with a Thai board-certified physician. Given the generated SOAP note and reference doctor--patient conversation, the physician independently evaluated each generated SOAP statement by assigning a binary factual grounding label using the same evaluation protocol as the LLM judge.
Table~\ref{tab:judge_alignment} demonstrates a Cohen's $\kappa$ score of 0.696, indicating substantial agreement with the physician's annotations.

\begin{figure*}[t]
\centering
\begin{promptbox}{SOAPPrecision Prompt}
\promptsize
Please act as an impartial clinical grounding judge. Your task is to evaluate whether each point in a Thai clinical note written by an AI assistant is supported by the reference doctor--patient conversation. For each point, output:\\
\texttt{factual\_prediction = 1} if the whole point is supported by the conversation, either directly, by clear paraphrase, or by a strong and conventional SOAP-style clinical inference.\\
\texttt{factual\_prediction = 0} if any clinically meaningful part of the point is unsupported, contradicted, over-specific, ambiguous, or only medically plausible.

This is a grounding task, not a medical correctness task. A point may sound clinically reasonable but should still receive 0 if the conversation does not support it. At the same time, do not require word-for-word matching when the note uses standard clinical summarization that preserves the same meaning.

\smallskip
\textbf{Judging procedure:}

1. Break the point into clinically meaningful subclaims. Examples include: symptoms/problems; diagnoses/assessments; disease status or severity; laboratory values and vital signs; medications and treatments; dose, timing, duration, frequency; body location, laterality, or qualifiers; causes, purposes, or rationales; demographic details; negative findings.

2. Evaluate each clinically meaningful subclaim separately.

3. Output 1 only if all clinically meaningful subclaims are supported.

4. Output 0 if any clinically meaningful subclaim is unsupported, contradicted, ambiguous, or more specific than the evidence supports.

5. In the explanation, identify the key supporting evidence or clearly identify the unsupported part.

\smallskip
\textbf{Allowed support:}
Direct statements or close paraphrases; Evidence combined across multiple parts of the conversation; Conventional SOAP summarization that preserves the same clinical meaning; Clinical normalization of shorthand, abbreviations, informal language, or ASR-noisy wording when the intended meaning is clear from context; Strong assessment-level inference supported by multiple aligned pieces of evidence from the history, examination, investigations, physician discussion, and plan. The goal is to judge clinical meaning, not exact wording.

\smallskip
\textbf{Clinical summarization rules:}
Accept conventional SOAP phrasing when it does not introduce new clinical facts; Accept standard clinical terminology that directly describes findings already present in the conversation; Accept common assessment labels when the overall clinical context strongly supports them; Accept routine plan summaries when they accurately describe the action discussed and do not add new details; Do not reject a point solely because the note is more concise or more clinically phrased than the conversation.

\smallskip
\textbf{Thai language and shorthand rules:}
Thai clinical shorthand may be normalized only when both the intended concept and value are clear from context; Do not convert shorthand into a more specific laboratory test, diagnosis, or measurement when multiple interpretations remain plausible; If the conversation clearly supports only a broader concept, do not automatically infer a more specific clinical variable; If ASR wording is noisy but the intended meaning is clear, accept it; If multiple interpretations remain plausible, treat over-specific interpretations as unsupported.

\smallskip
\textbf{Demographic rules:}
Sex/gender can be accepted when explicitly stated or when the patient's own speech consistently and coherently indicates it with clear speaker attribution; Do not infer sex/gender from isolated cues, from the physician's speech, or when speaker identity is ambiguous; Do not infer nationality, age, occupation, or other demographic information unless supported by the conversation.

\smallskip
\textbf{Medication rules:}
Medication continuation, discontinuation, modification, exact drug names, dosage, timing, frequency, and treatment purpose are high-risk details and require strong support; Do not infer medication continuation merely because no stop order was given; Do not infer medication continuation from medication history, adherence discussion, or medication lists alone; Do not infer exact medication details unless they are explicitly stated or unambiguously derivable from stated information; Accept medication class normalization only when the intended class is clear from the conversation.

\smallskip
\textbf{Causality and purpose rules:}
Do not create causal, explanatory, or purpose relationships simply because two facts appear together; Accept causality or purpose only when the relationship is explicitly stated or clearly discussed; Reject explanations, rationales, treatment purposes, or referral reasons that are merely medically plausible but not grounded in the conversation.

\smallskip
\textbf{Disease status rules:}
Qualifiers such as controlled, uncontrolled, stable, progressive, improving, worsening, resolved, or disease stage require stronger evidence than the diagnosis itself; Do not infer disease status from a single isolated clue; Accept disease status only when supported by trends, targets, comparisons, physician assessment, management decisions, or multiple converging pieces of evidence.

\smallskip
\textbf{Numeric rules:}
Exact values must be supported by the conversation; Approximate wording may support approximate or range-based statements, but not precise values; Do not convert approximate values into exact measurements; Do not add precision that is absent from the conversation.

\smallskip
\textbf{Negative finding rules:}
Do not infer absence of symptoms, absence of disease, absence of investigations, absence of risk factors, or absence of history from silence; Accept negative findings only when explicitly stated, directly answered, or clearly communicated; For bundled negative findings, each negative finding must be supported.

\smallskip
\textbf{Bundled point rules:}
Judge the whole point; Every clinically meaningful component must be supported; However, do not treat routine restatements of clearly described clinical actions as new unsupported facts; New diagnoses, medications, tests, causes, purposes, severities, timelines, disease-status qualifiers, or other meaningful additions require support.

\smallskip
\textbf{Follow-up rules:}
Follow-up intervals must be supported by the conversation; Follow-up purposes must be explicitly stated or strongly tied to the physician's discussion and plan; Do not add future investigations, monitoring goals, or follow-up reasons that were not discussed.

\smallskip
\textbf{Decision calibration:}
Be strict for high-risk factual details such as medications, doses, exact values, causality, disease status, future plans, and treatment rationale; Be flexible for standard SOAP summarization, common clinical terminology, and clearly grounded shorthand; Under uncertainty about a clinically meaningful detail, output 0; If \texttt{factual\_prediction} is 0, explicitly identify the unsupported, contradicted, ambiguous, or over-specific component; If the explanation concludes that the point is clearly supported, \texttt{factual\_prediction} should normally be 1.

\smallskip
Return structured JSON with one evaluation for every point, preserving each original \texttt{point\_id} and point text.
\end{promptbox}
\caption{Prompt used for the \emph{SOAPPrecision} judge. Given a generated SOAP note and the encounter transcript, the judge decides whether each clinical statement in the note is factually supported by the transcript. The worked example included in the prompt is omitted here.}
\label{fig:spprompt}
\end{figure*}

\clearpage

\section{Additional Training Details}
\label{app:trainings}

All experiments were conducted on one NVIDIA H100 NVL GPU.
For ROUGE-L, we use PyThaiNLP's newmm tokenizer~\cite{pythai}. For BLEURT, we use bleurt-oss-21/BLEURT-20.

\paragraph{Supervised Fine-tuning.}
We fine-tune Gemma-4-E4B-it~\cite{gemmateam2026gemma4technicalreport} using supervised fine-tuning (SFT) with LoRA~\cite{lora} ($r=32$, $\alpha=64$) using the Unsloth framework~\citep{unsloth}. Training uses a maximum sequence length of 8,192, a learning rate of $9\times10^{-5}$ with a cosine decay scheduler, AdamW~\cite{adamw} optimizer, batch size of 16, and 6 epochs, and selects the checkpoint with the lowest validation loss.

\paragraph{GRPO.}
We train GRPO on Gemma-4-E4B-it~\cite{gemmateam2026gemma4technicalreport} with LoRA~\cite{lora} ($r=32$, $\alpha=64$) on the Unsloth framework using a learning rate of $8\times10^{-6}$ and the AdamW optimization with a cosine decay scheduler. Training uses a batch size of 16 with 8 sampled generations per prompt with 0.4 temperature, a maximum completion length of 4,096 tokens, and 300 optimization steps. All LLM judges used as reward functions in the GRPO training are evaluated by Grok-4.3 using the same instructions to lower training costs.
For the weights of each reward presented in Section~\ref{sec:rlvr}, we use $\lambda_{\mathrm{format}} = 3$, $\lambda_{\mathrm{AF}} = 2$, $\lambda_{\mathrm{sig}} = 1$, $\lambda_{\mathrm{comp}} = 4$, and $\lambda_{\mathrm{fact}} = 3$ as the weights of $R_{\mathrm{format}}$, $R_{\mathrm{AF}}$, $R_{\mathrm{sig}}$, $R_{\mathrm{comp}}$, and $R_{\mathrm{fact}}$, respectively. The final reward is computed as their weighted sum:
\begin{equation}
\begin{aligned}
R_\text{all} ={}&
\lambda_{\mathrm{format}} R_{\mathrm{format}}
+ \lambda_{\mathrm{AF}} R_{\mathrm{AF}}
+ \lambda_{\mathrm{sig}} R_{\mathrm{sig}} \\
&+ \lambda_{\mathrm{comp}} R_{\mathrm{comp}}
+ \lambda_{\mathrm{fact}} R_{\mathrm{fact}}.
\end{aligned}
\end{equation}

These weights are used for all ASCRIBE-based GRPO models.

For $R_{\mathrm{sig}}$, the Atomic Fact LLM Judge may match a generated fact to multiple reference facts. Let $C(i)=\{c(i)_1,\ldots\}$ denote the set of reference fact indices matched to the generated fact $f_i$. The predicted significance label $s_i$ is considered correct if it matches the significance label of any matched reference fact. Otherwise, the prediction is considered incorrect.

\section{Additional Details on ASR Robustness Experiments}
\label{app:asr}
In Figure~\ref{fig:asr}, we compare LLM-judge metrics on manually transcribed and ASR transcripts to evaluate robustness to transcription noise. ASR transcripts are generated using a proprietary Whisper-based model~\cite{aung-etal-2024-thonburian} fine-tuned on an internal voicebank containing Thai clinical speech, which obtains a WER of 29.22 on the ThaiClinicBench dataset.

\section{Qualitative Results}
\label{app:qualitative}

We present two representative encounters to examine how structured reasoning affects the clinical information preserved in the generated SOAP notes. For each example, we report the generated notes and provided a fact-level analysis showing which facts are entailed. All English translations are attached for readability and are made by Claude Opus.

In the first example (Tables~\ref{tab:qualitative_soap_notes} and~\ref{tab:qualitative_claim_entailment}), all models capture the main symptoms, including dizziness, nausea, and headache. However, the structured prompt with Gemini 3.1 Pro preserves more contextual details, such as working with chemicals throughout the day and normal limb strength, achieving 8/11 entailed facts compared with 6/11 for free-form CoT. The GRPO model trained on real encounters also captures normal limb strength and the lack of nasal protection, but omits the two-day symptom duration and full-day chemical exposure, yielding 7/11 entailed facts.

The second example
(Tables~\ref{tab:qualitative_grpo_transcript},~\ref{tab:qualitative_grpo_soap_notes} and~\ref{tab:qualitative_grpo_claim_entailment}) compares different GRPO reward and training-data settings. All three models recover the major symptoms, examination findings, viral assessment, and symptomatic treatment. Nevertheless, the model trained with our rewards on real encounters retains the most complete set of details, including pain during phonation and the instruction to monitor for tonsillar swelling or exudate after returning home, achieving 22/28 entailed facts. The synthetic-only model performs similarly on 21/28 facts but misses several contextual details, whereas the \textsc{ClinicEval} model omits more follow-up- and encounter-specific information, achieving 17/28 facts.

\begin{table*}[t]
\centering
\setlength{\tabcolsep}{4pt}
\renewcommand{\arraystretch}{1.1}
\scriptsize
\begin{tabularx}{\textwidth}{
    >{\raggedright\arraybackslash}p{0.065\textwidth}
    >{\raggedright\arraybackslash\scriptsize}X
    >{\raggedright\arraybackslash\scriptsize}X
    >{\raggedright\arraybackslash\scriptsize}X
}
\toprule
\multicolumn{4}{l}{\small\textbf{Source transcript}} \\
\midrule
\multicolumn{4}{
    >{\raggedright\arraybackslash\scriptsize}
    p{\dimexpr\textwidth-2\tabcolsep\relax}
}{%
    \textit{\thaicell{%
        เออ มีอยู่โอเค เมื่อก่อนมันเป็น queue นัดหมาย
        มันไม่มีเสียงแบบว่า รอ มีนัดเหรอครับ
        สวัสดีค่ะ สวัสดีครับ เป็นอะไร มาหรอครับ
        อืมมันเวียนหัวแล้วก็คลื่นไส้ เพราะว่า ประมาณ
        เป็นน้ำแอลกอฮอล์หรืออะไรสักอย่าง อ่า
        แต่เราทำงานเราไม่ได้ปิดจมูก
        อ๋อ ทำโรงงาน แบบ สารเคมีแล้วเวียนศีรษะ
        ใช่ ๆ ๆ ที่ทำเนี้ยทั้งวันเลยวันแรกเค้าให้พี่มาช่วย
        อ๋อ ปวดหัวไหม ปวดหัว แล้วก็คือ 2 วันแล้ว
        ข้างเดียวหรือ 2 ข้าง ข้างเดียว
        ข้างไหน ข้าง ข้าง ๆ นี้ ขวา ใช่
        อ่อ แขนขา มีแรงปกติดี มีแรงปกติ}}%
    \par\vspace{3pt}
    \textit{%
        Uh, yes, okay. It used to be an appointment queue.
        There was no sound, like, wait, do you have an appointment?
        Hello. Hello. What brings you in?
        Um, I feel dizzy and nauseous, because, roughly,
        it's alcohol or something, ah.
        But when we work, we don't cover our nose.
        Oh, you work in a factory, with chemicals, and then got dizzy.
        Yes, yes, yes, doing this all day; on the first day they had me come help.
        Oh, any headache? Headache, and it's been 2 days.
        One side or both sides? One side.
        Which side? This side, right here. Right. Yes.
        Oh, arms and legs have normal strength. Normal strength.}%
} \\
\midrule
&
\small\textbf{\makecell[l]{Gemini 3.1 Pro\\CoT}} &
\small\textbf{\makecell[l]{Gemini 3.1 Pro\\Ours}} &
\small\textbf{\makecell[l]{GRPO Gemma-4-E4B\\Ours (Real)}} \\
\midrule
\small\textbf{CE} &
\small 54.5 (6/11 claims) &
\small 72.7 (8/11 claims) &
\small 63.6 (7/11 claims) \\
\midrule
\small\textbf{HPI} &
\thaicell{%
    ผู้ป่วยมีอาการเวียนศีรษะ คลื่นไส้ และปวดศีรษะข้างขวา เป็นมา 2 วัน;
    มีประวัติทำงานในโรงงานและสูดดมสารเคมี (คล้ายแอลกอฮอล์) ตลอดทั้งวันโดยไม่ได้ใส่หน้ากากป้องกัน;
    ปฏิเสธอาการแขนขาอ่อนแรง.} &
\thaicell{%
    ผู้ป่วยมีอาการเวียนศีรษะ คลื่นไส้ และปวดศีรษะข้างขวาข้างเดียว เป็นมา 2 วัน;
    ผู้ป่วยสงสัยว่าอาการเกิดจากการสูดดมสารเคมี (แอลกอฮอล์หรือสารอื่น) จากการทำงานในโรงงาน;
    ผู้ป่วยทำงานสัมผัสสารเคมีทั้งวันในวันแรกที่ไปช่วยงาน โดยไม่ได้ใส่หน้ากากป้องกัน;
    แขนขามีแรงปกติดี.} &
\thaicell{%
    มีอาการเวียนหัวและคลื่นไส้;
    อาการอาจเกี่ยวข้องกับน้ำแอลกอฮอล์หรือสารบางอย่าง;
    ผู้ป่วยทำงานในโรงงานที่เกี่ยวข้องกับสารเคมีและไม่ได้ปิดจมูกขณะทำงาน;
    มีอาการเวียนศีรษะขณะทำงานทั้งวัน;
    มีอาการปวดหัวเป็นมา 2 วันแล้ว;
    แขนขาไม่มีแรงปกติ.} \\
\cmidrule[0.1pt](l){2-4}
&
\entext{%
    The patient has dizziness, nausea, and a right-sided headache for 2 days;
    has a history of working in a factory and inhaling chemicals (alcohol-like) all day without wearing a protective mask;
    denies limb weakness.} &
\entext{%
    The patient has dizziness, nausea, and a one-sided right headache for 2 days;
    the patient suspects the symptoms are caused by inhaling chemicals (alcohol or another substance) from working in a factory;
    the patient worked with chemical exposure all day on the first day of helping out, without wearing a protective mask;
    limb strength is normal.} &
\entext{%
    Has dizziness and nausea;
    the symptoms may be related to alcohol or some substance;
    the patient works in a factory involving chemicals and does not cover the nose while working;
    has dizziness while working all day;
    has had a headache for 2 days;
    limbs do not have normal strength.} \\
\midrule
\small\textbf{PE} &
\entext{--} &
\entext{Motor power: normal.} &
\thaicell{แขนขามีแรงปกติ.} \\
\cmidrule[0.1pt](l){2-4}
&
\entext{--} &
\entext{Motor power: normal.} &
\entext{Limbs have normal strength.} \\
\midrule
\small\textbf{Results} &
\entext{--} &
\entext{--} &
\entext{--} \\
\cmidrule[0.1pt](l){2-4}
&
\entext{--} &
\entext{--} &
\entext{--} \\
\midrule
\small\textbf{A\&P} &
\entext{--} &
\entext{Headache and dizziness, likely secondary to chemical exposure.} &
\entext{--} \\
\cmidrule[0.1pt](l){2-4}
&
\entext{--} &
\entext{Headache and dizziness, likely secondary to chemical exposure.} &
\entext{--} \\
\bottomrule
\end{tabularx}
\caption{
    Qualitative comparison of SOAP notes generated by different methods from the same source
    transcript. CE is the ClaimEntailment score, computed against the same set of 11
    transcript-derived reference claims. To preserve space, statements are separated with semi-colons ($;$). Within each section, the top half is the Thai note as generated and the bottom half is a literal English translation.
}
\label{tab:qualitative_soap_notes}
\end{table*}


\begin{table*}[t]
\centering
\setlength{\tabcolsep}{4pt}
\renewcommand{\arraystretch}{1.8}
\scriptsize
\begin{tabularx}{\textwidth}{
    >{\centering\arraybackslash\small}p{0.03\textwidth}
    >{\raggedright\arraybackslash\linespread{1.2}\scriptsize}X
    >{\centering\arraybackslash\small}p{0.08\textwidth}
    >{\centering\arraybackslash\small}p{0.08\textwidth}
    >{\centering\arraybackslash\small}p{0.08\textwidth}
}
\toprule
\textbf{\#} &
\small\textbf{Reference claim} &
\makecell[c]{\textbf{Gemini}\\\textbf{CoT}} &
\makecell[c]{\textbf{Gemini}\\\textbf{Ours}} &
\makecell[c]{\textbf{Gemma}\\\textbf{Ours}} \\
\midrule
1 &
\thaitext{ผู้ป่วยมีอาการเวียนหัว} \entext{(The patient has dizziness)} &
\cmark & \cmark & \cmark \\
2 &
\thaitext{ผู้ป่วยมีอาการคลื่นไส้} \entext{(The patient has nausea)} &
\cmark & \cmark & \cmark \\
3 &
\thaitext{ ผู้ป่วยระบุว่าอาการเกิดจากการทำงาน กับสารเคมีคล้ายน้ำแอลกอฮอล์ } \entext{(The patient states the symptoms came from working with alcohol-like chemicals)} &
\xmark & \xmark & \xmark \\
4 &
\thaitext{ผู้ป่วยไม่ได้ปิดจมูกขณะทำงาน} \entext{(The patient did not cover their nose while working)} &
\xmark & \xmark & \cmark \\
5 &
\thaitext{ผู้ป่วยทำงานในโรงงาน} \entext{(The patient works in a factory)} &
\cmark & \cmark & \cmark \\
6 &
\thaitext{ผู้ป่วยทำงานกับสารเคมีทั้งวัน} \entext{(The patient works with chemicals all day)} &
\xmark & \cmark & \xmark \\
7 &
\thaitext{ วันที่เกิดอาการเป็นวันแรก ที่ผู้ป่วยถูกให้มาช่วยทำงาน } \entext{(The day symptoms began was the patient's first day being brought in to help)} &
\xmark & \xmark & \xmark \\
8 &
\thaitext{ผู้ป่วยมีอาการปวดหัว} \entext{(The patient has a headache)} &
\cmark & \cmark & \cmark \\
9 &
\thaitext{ผู้ป่วยมีอาการมา 2 วันแล้ว} \entext{(The patient has had symptoms for 2 days)} &
\cmark & \cmark & \xmark \\
10 &
\thaitext{ผู้ป่วยปวดหัวข้างขวา} \entext{(The patient has a right-sided headache)} &
\cmark & \cmark & \cmark \\
11 &
\thaitext{ผู้ป่วยมีแรงแขนขาปกติ} \entext{(The patient has normal limb strength)} &
\xmark & \cmark & \cmark \\
\midrule
\multicolumn{2}{r}{\small\textbf{Total entailed claims}} &
\textbf{6/11} & \textbf{8/11} & \textbf{7/11} \\
\multicolumn{2}{r}{\small\textbf{ClaimEntailment score}} &
\textbf{54.5} & \textbf{72.7} & \textbf{63.6} \\
\bottomrule
\end{tabularx}
\caption{
    Claim-level entailment comparison of the generated SOAP notes in
    Table~\ref{tab:qualitative_soap_notes}. \cmark{} indicates that the reference claim is entailed
    by the generated SOAP note under the ClaimEntailment metric, whereas \xmark{} indicates that it
    is not fully entailed. Scores are reported as percentages.
}
\label{tab:qualitative_claim_entailment}
\end{table*}


\begin{table*}[t]
\centering
\setlength{\tabcolsep}{4pt}
\renewcommand{\arraystretch}{1.1}
\scriptsize
\begin{tabularx}{\textwidth}{>{\raggedright\arraybackslash\scriptsize}X}
\toprule
\small\textbf{Source transcript} \\
\midrule
    \textit{\thaicell{%
        หวัดดีครับ หวัดดีค่ะ วันนี้เป็นไรมาเอ่ย เจ็บคอค่ะ
        ครับ ไปทำอะไรมาแล้วเจ็บคอ อืม แบบไปเที่ยวไปเจอคนอะไรงี้มีไหม
        ไม่ ๆ ไม่มีค่ะ ไม่ ที่ทำงานมีใครไม่สบายไหม ไม่ มีค่ะ ไม่มีเลยเนาะ
        เป็นมากี่วันแล้วครับ เมื่อวาน เมื่อวานเนาะ กลืนน้ำลายได้ไหม
        กลืนน้ำลายแล้วเจ็บเลยไหม กลืน ๆ ได้แต่ว่าเวลากินน้ำก็จะรู้สึกเจ็บนิดนึง
        อืม มีไข้ไหมครับ ไม่มีค่ะ ไม่มีไข้เนาะ ขอดูคอนิดนึงนะ อ่า ทำ
        อา แป๊บนะ ทำเสียงอา อา เจ็บเนาะเหมือนแบบมันสะอึก
        อะเงยหน้าขึ้นนิดนึง โอเค โอเค คอแดงนิดนึงนะ
        แล้วก็ แต่คอไม่มีหนองนะครับ มี ไอไหม มีนิดนึง
        มีปวดเมื่อยตามตัวไหมครับ ปวด ปวดเมื่อยตามตัวใช่ไหมครับ
        อ๋อ อ๋อ แต่ไอ ๆ ด้วยนิดนึงเนาะ มีค่ะ
        มีไข้มีไหมที่บ้านดูตัวรุม ๆ มีไหม ไม่ ไม่มีเลยเนาะ
        โอเค เอ่อ โอเคเดี๋ยวผมให้ยาไปนะ นะครับ
        มีไอเนี่ยมีเสมหะหรือไอแห้ง ๆ ไอแห้ง ๆ ไอแห้ง ๆ เนาะ
        แต่ไอเยอะจนรำคาญเลยไหมหรือว่าแบบไม่ได้เยอะ
        ไม่ค่อยมีนิดเดียว นิดเดียวเลยเนาะ นิดเดียว
        แต่ว่าที่ดูคอแล้ว คอไม่มี ไม่มีหนองในคอนะ
        แต่ว่าแนะนำว่าถ้ากลับบ้านไปสังเกตหน้าปากดูคอได้ว่า
        ตรง tonsil ข้าง ๆ ในคอมันมีบวมมี ๆ หนองบ้างไหม
        คือบางคนเนี่ยพอมันอาการเป็นวันแรกอะ
        มันจะแยกยากว่าเป็น bacteria หรือเป็น virus นะครับ
        แต่จากอาการเนี่ยค่อนข้างเป็น virus
        โทษนะ ขอคลำต่อมน้ำเหลืองนิดนึงนะ
        โอเค คลำต่อมน้ำเหลืองบริเวณคอนะครับ
        cervical lymph node เนี่ยไม่มีนะครับ
        ไม่มีต่อมน้ำเหลืองที่คอโตเนาะ
        ซึ่งก็อาจจะมีแนวโน้มเป็นอาการติดเชื้อ virus ได้มากกว่า
        ซึ่งพอเป็นการติดเชื้อ virus เนี่ย
        เราจะแนะนำให้กินยารักษาตามอาการแล้วพักผ่อนเยอะ ๆ
        เอาใบลาไหม ลากี่วันดีวันนี้พรุ่งนี้
        วันเดียวพอ พรุ่งนี้ วันเดียวพรุ่งนี้ พรุ่งนี้ พรุ่งนี้โอเค
        ได้เดี๋ยวหมอให้เอายา ยาหยุดนี้ไปด้วยนะ
        หมายถึงว่าตามปกตินั่นแหละ นั่นแหละเนาะ นะครับ
        โอเค ได้ มีข้อสงสัยอยากสอบถามตอนไหนเพิ่มเติมไหมครับ
        ไม่ ไม่มีเลยเนาะ เดี๋ยวลองกินยานะ
        โอเคค่ะขอบคุณมากค่ะ ครับหวัดดีครับ}}%
    \par\vspace{3pt}
    \textit{%
        Hello. Hello. What brings you in today? A sore throat.
        Okay. What were you doing before the sore throat? Um, like, did you travel or meet anyone?
        No, no, I didn't. No. Is anyone at work sick? No, there isn't. No one at all, right?
        How many days has it been? Since yesterday. Yesterday, right. Can you swallow saliva?
        Does it hurt when you swallow? I can swallow, but when I drink water it hurts a little.
        Um, any fever? No. No fever, right. Let me take a look at your throat. Ah, say ah.
        Hold on, say ``ah''. Ah. It hurts, right, like a hiccup.
        Tilt your head up a little. Okay, okay, your throat is a little red,
        and, but there's no pus in your throat. Any cough? A little.
        Any body aches? Yes. Body aches, right?
        Oh, oh, and a bit of a cough too, right? Yes.
        Any fever, feeling feverish at home? No. None at all, right.
        Okay, uh, okay, I'll give you some medicine.
        Is the cough with phlegm, or dry? Dry. Dry cough, right.
        But is the cough so bad that it bothers you, or not much?
        Not much, just a little. Just a little, right. Just a little.
        But from looking at your throat, there's no pus in your throat.
        But I recommend that when you get home, you look at your throat in front of the mirror to see
        whether the tonsils at the sides of your throat are swollen or have any pus.
        For some people, on the first day of symptoms,
        it's hard to tell whether it's bacteria or a virus.
        But from the symptoms, it's likely a virus.
        Excuse me, let me feel your lymph nodes.
        Okay, palpating the lymph nodes in the neck,
        no cervical lymph nodes.
        No enlarged lymph nodes in the neck, right.
        So it's more likely to be a viral infection.
        And with a viral infection,
        we recommend taking medicine for the symptoms and getting plenty of rest.
        Do you want a leave note? How many days, today or tomorrow?
        One day is enough, tomorrow. One day, tomorrow. Tomorrow. Tomorrow, okay.
        Sure, I'll have you take the medicine, and this ``stop medicine'' too.
        I mean, the usual. Right, okay.
        Okay, sure. Any other questions you'd like to ask?
        No, none at all, right. Try taking the medicine.
        Okay, thank you very much. Bye.}%
    \\
\bottomrule
\end{tabularx}
\caption{
    Source transcript of the second qualitative example, used for the SOAP notes in
    Table~\ref{tab:qualitative_grpo_soap_notes}. The top half is the Thai transcript and the
    bottom half is a literal English translation.
}
\label{tab:qualitative_grpo_transcript}
\end{table*}


\begin{table*}[tp]
\centering
\setlength{\tabcolsep}{4pt}
\renewcommand{\arraystretch}{1.1}
\scriptsize
\begin{tabularx}{\textwidth}{
    >{\raggedright\arraybackslash}p{0.065\textwidth}
    >{\raggedright\arraybackslash\scriptsize}X
    >{\raggedright\arraybackslash\scriptsize}X
    >{\raggedright\arraybackslash\scriptsize}X
}
\toprule
&
\small\textbf{ClinicEval} &
\small\textbf{Ours (Real)} &
\small\textbf{Ours (Synth)} \\
\midrule
\small\textbf{CE} &
\small 60.7 (17/28 claims) &
\small 78.6 (22/28 claims) &
\small 75.0 (21/28 claims) \\
\midrule
\small\textbf{HPI} &
\thaicell{%
    ผู้ป่วยมาด้วยอาการเจ็บคอ (Sore throat);
    ไม่ทราบว่าไปเที่ยวหรือเจอคนที่มีอาการหรือไม่
    (No recent travel or exposure to sick people);
    ไม่พบผู้ป่วยไม่สบายที่ที่ทำงาน
    (No sick colleagues at work);
    อาการเป็นมาตั้งแต่เมื่อวาน
    (Symptoms started yesterday);
    กลืนน้ำลายได้ แต่เจ็บเล็กน้อยเวลากินน้ำ
    (Can swallow saliva, but feels slight pain when drinking water);
    ไม่มีไข้ (No fever);
    มีอาการไอเล็กน้อย (Mild cough);
    มีอาการปวดเมื่อยตามตัว (Body aches);
    ไม่มีไข้ที่บ้าน (No fever at home);
    ลักษณะไอเป็นไอแห้ง (Dry cough), ไอไม่เยอะมาก (Not much cough);
    จากการตรวจคอ ไม่พบหนองในคอ
    (No pus observed in the throat);
    ผู้ป่วยไม่แน่ใจว่าไอแห้งหรือมีเสมหะ
    แต่ระบุว่าไอแห้ง
    (Patient unsure if phlegm or dry cough, but stated dry cough).} &
\thaicell{%
    ผู้ป่วยมาด้วยอาการเจ็บคอ;
    อาการเริ่มเป็นเมื่อวาน;
    ไม่เคยไปเที่ยวหรือเจอคนป่วย;
    ไม่มีคนในที่ทำงานป่วย;
    กลืนน้ำลายได้แต่เจ็บเล็กน้อยเวลากินน้ำ;
    ไม่มีไข้;
    ทำเสียงอาแล้วเจ็บเหมือนสะอึก;
    มีไอเล็กน้อย;
    มีอาการปวดเมื่อยตามตัว;
    ไม่มีไข้ที่บ้าน;
    ไอเล็กน้อย ไม่ค่อยมี;
    คอไม่มีหนอง.} &
\thaicell{%
    ผู้ป่วยมาด้วยอาการเจ็บคอ;
    อาการเป็นมาตั้งแต่เมื่อวาน;
    ผู้ป่วยไม่ได้ไปเที่ยวหรือเจอคนอื่นที่ทำให้เจ็บคอ;
    ที่ทำงานไม่มีใครไม่สบาย;
    กลืนน้ำลายได้แต่เจ็บ;
    เวลากินน้ำรู้สึกเจ็บนิดนึง;
    ไม่มีไข้;
    มีไอเล็กน้อย;
    มีปวดเมื่อยตามตัว;
    ที่บ้านดูตัวรุม ๆ ไม่มี;
    ไอแห้ง ๆ;
    ไอไม่เยอะ นิดเดียว.} \\
\cmidrule[0.1pt](l){2-4}
&
\entext{%
    The patient presents with a sore throat;
    unknown whether they traveled or met anyone with symptoms;
    no sick people found at work;
    symptoms since yesterday;
    can swallow saliva, but slight pain when drinking water;
    no fever;
    mild cough;
    body aches;
    no fever at home;
    the cough is dry, not much coughing;
    on throat examination, no pus found in the throat;
    the patient is unsure whether the cough is dry or productive, but states it is dry.} &
\entext{%
    The patient presents with a sore throat;
    symptoms started yesterday;
    has not traveled or met sick people;
    no one at work is sick;
    can swallow saliva but has slight pain when drinking water;
    no fever;
    saying ``ah'' hurts, like a hiccup;
    mild cough;
    body aches;
    no fever at home;
    slight cough, not much;
    no pus in the throat.} &
\entext{%
    The patient presents with a sore throat;
    symptoms since yesterday;
    the patient did not travel or meet others who caused the sore throat;
    no one at work is sick;
    can swallow saliva but it hurts;
    feels slight pain when drinking water;
    no fever;
    mild cough;
    body aches;
    no feeling feverish at home;
    dry cough;
    not much cough, just a little.} \\
\midrule
\small\textbf{PE} &
\thaicell{%
    ตรวจคอ: คอแดงเล็กน้อย (Throat: Mild redness);
    ตรวจคอ: ไม่พบหนองในคอ (Throat: No pus);
    คลำต่อมน้ำเหลืองบริเวณคอ:
    ไม่พบต่อมน้ำเหลืองโต
    (Cervical lymph node palpation:
    No enlarged cervical lymph nodes).} &
\thaicell{%
    ตรวจคอ: คอแดงเล็กน้อย;
    ตรวจคอ: ไม่มีหนองในคอ;
    คลำต่อมน้ำเหลืองบริเวณคอ
    (cervical lymph node):
    ไม่มีต่อมน้ำเหลืองโต.} &
\thaicell{%
    ตรวจคอพบคอแดงนิดนึง;
    คอไม่มีหนอง;
    คลำต่อมน้ำเหลืองบริเวณคอ
    cervical lymph node ไม่มีโต.} \\
\cmidrule[0.1pt](l){2-4}
&
\entext{%
    Throat exam: mild redness;
    throat exam: no pus found in the throat;
    neck lymph node palpation: no enlarged lymph nodes.} &
\entext{%
    Throat exam: mildly red;
    throat exam: no pus in the throat;
    neck lymph node (cervical lymph node) palpation: no enlarged lymph nodes.} &
\entext{%
    Throat exam shows slight redness;
    no pus in the throat;
    palpation of the neck lymph nodes: cervical lymph nodes not enlarged.} \\
\midrule
\small\textbf{Results} &
\thaicell{%
    จากอาการและผลการตรวจ
    ค่อนข้างมีแนวโน้มเป็น Viral infection
    มากกว่า Bacterial infection;
    แนะนำให้สังเกตอาการที่หน้าปากคอว่า
    มี tonsil บวมหรือมีหนองหรือไม่.} &
\thaicell{%
    จากอาการและผลการตรวจ ค่อนข้างเป็น virus;
    แนวโน้มเป็นอาการติดเชื้อ virus มากกว่า.} &
\thaicell{%
    อาการค่อนข้างเป็น virus;
    มีแนวโน้มเป็นอาการติดเชื้อ virus มากกว่า.} \\
\cmidrule[0.1pt](l){2-4}
&
\entext{%
    Based on the symptoms and examination, more likely a viral infection than a bacterial infection;
    advised to check in front of the mouth and throat whether the tonsils are swollen or have pus.} &
\entext{%
    Based on the symptoms and examination, likely a virus;
    more likely a viral infection.} &
\entext{%
    Symptoms likely viral;
    more likely a viral infection.} \\
\midrule
\small\textbf{A\&P} &
\thaicell{%
    วินิจฉัยเบื้องต้น: น่าจะเป็น Viral infection;
    แนะนำให้รับประทานยาตามอาการ;
    พักผ่อนให้มาก;
    แนะนำให้ลาหยุด 1 วัน;
    ให้ยาตามปกติ;
    หากมีข้อสงสัยให้สอบถามเพิ่มเติม.} &
\thaicell{%
    วินิจฉัย: คาดว่าติดเชื้อไวรัส
    (Viral infection);
    แนะนำให้กินยารักษาตามอาการและพักผ่อนเยอะ ๆ;
    แนะนำให้สังเกตหน้าปากดูคอว่า
    มี tonsil บวมหรือมีหนองหรือไม่เมื่อกลับบ้าน;
    ให้ใบลา 1 วัน (พรุ่งนี้);
    ให้ยาตามปกติ.} &
\thaicell{%
    แนะนำให้สังเกตหน้าปากดูคอว่า
    tonsil ข้าง ๆ ในคอมันมีบวมมีหนองบ้างไหม;
    แนะนำให้กินยารักษาตามอาการและพักผ่อนเยอะ ๆ;
    ให้ใบลา 1 วัน (พรุ่งนี้);
    ให้ยาหยุดนี้ไปด้วย.} \\
\cmidrule[0.1pt](l){2-4}
&
\entext{%
    Preliminary diagnosis: likely a viral infection;
    advised to take symptomatic medication;
    rest a lot;
    advised to take 1 day of leave;
    medication given as usual;
    ask if there are further questions.} &
\entext{%
    Diagnosis: suspected viral infection (Viral infection);
    advised to take symptomatic medication and rest a lot;
    advised to check the throat in front of the mouth for swollen tonsils or pus after returning home;
    1-day leave note given (tomorrow);
    medication given as usual.} &
\entext{%
    Advised to check the throat in front of the mouth for whether the tonsils at the sides are swollen or have pus;
    advised to take symptomatic medication and rest a lot;
    1-day leave note given (tomorrow);
    also given this ``stop medicine''.} \\
\bottomrule
\end{tabularx}
\caption{
    Qualitative comparison of SOAP notes generated by three GRPO models from the source
    transcript in Table~\ref{tab:qualitative_grpo_transcript}. CE is the ClaimEntailment score,
    computed against the same set of 28 transcript-derived reference claims. To preserve space,
    statements are separated with semi-colons ($;$). Within each section, the top half is the
    Thai note as generated (English in parentheses was generated by the model) and the bottom
    half is a literal English translation.
}
\label{tab:qualitative_grpo_soap_notes}
\end{table*}


\begin{table*}[t]
\centering
\setlength{\tabcolsep}{4pt}
\renewcommand{\arraystretch}{1.8}
\scriptsize
\begin{tabularx}{\textwidth}{
    >{\centering\arraybackslash\small}p{0.03\textwidth}
    >{\raggedright\arraybackslash\linespread{1.2}\scriptsize}X
    >{\centering\arraybackslash\small}p{0.08\textwidth}
    >{\centering\arraybackslash\small}p{0.08\textwidth}
    >{\centering\arraybackslash\small}p{0.08\textwidth}
}
\toprule
\textbf{\#} &
\small\textbf{Reference claim} &
\makecell[c]{\textbf{Clinic}\\\textbf{Eval}} &
\makecell[c]{\textbf{Ours}\\\textbf{Real}} &
\makecell[c]{\textbf{Ours}\\\textbf{Synth}} \\
\midrule
1 &
\thaitext{ผู้ป่วยมีอาการเจ็บคอ} \entext{(The patient has a sore throat)} &
\cmark & \cmark & \cmark \\
2 &
\thaitext{ผู้ป่วยบอกว่าไม่ได้ไปเที่ยว} \entext{(The patient says they did not travel)} &
\xmark & \cmark & \cmark \\
3 &
\thaitext{ที่ทำงานของผู้ป่วยไม่มีคนไม่สบาย} \entext{(No one at the patient's workplace is sick)} &
\cmark & \cmark & \cmark \\
4 &
\thaitext{ผู้ป่วยเริ่มมีอาการเจ็บคอเมื่อวาน} \entext{(The patient's sore throat started yesterday)} &
\cmark & \cmark & \cmark \\
5 &
\thaitext{ผู้ป่วยกลืนน้ำลายได้} \entext{(The patient can swallow saliva)} &
\cmark & \cmark & \cmark \\
6 &
\thaitext{ผู้ป่วยรู้สึกเจ็บคอนิดนึงเวลากินน้ำ} \entext{(The patient feels slight throat pain when drinking water)} &
\cmark & \cmark & \cmark \\
7 &
\thaitext{ผู้ป่วยไม่มีไข้} \entext{(The patient has no fever)} &
\cmark & \cmark & \cmark \\
8 &
\thaitext{ผู้ป่วยรู้สึกเจ็บคอเมื่อทำเสียงอา} \entext{(The patient feels throat pain when saying ``ah'')} &
\xmark & \cmark & \xmark \\
9 &
\thaitext{ผู้ป่วยรู้สึกเหมือนสะอึกเมื่อทำเสียงอา} \entext{(The patient feels like hiccuping when saying ``ah'')} &
\xmark & \cmark & \xmark \\
10 &
\thaitext{แพทย์ตรวจพบคอแดงนิดนึง} \entext{(The physician finds slight throat redness)} &
\cmark & \cmark & \cmark \\
11 &
\thaitext{แพทย์ตรวจไม่พบคอมีหนอง} \entext{(The physician finds no pus in the throat)} &
\cmark & \cmark & \cmark \\
12 &
\thaitext{ผู้ป่วยมีอาการไอนิดนึง} \entext{(The patient has a slight cough)} &
\cmark & \cmark & \cmark \\
13 &
\thaitext{ผู้ป่วยมีอาการปวดเมื่อยตามตัว} \entext{(The patient has body aches)} &
\cmark & \cmark & \cmark \\
14 &
\thaitext{ผู้ป่วยมีอาการไอแห้ง} \entext{(The patient has a dry cough)} &
\cmark & \xmark & \cmark \\
15 &
\thaitext{ผู้ป่วยไอไม่เยอะ} \entext{(The patient does not cough much)} &
\cmark & \cmark & \cmark \\
16 &
\thaitext{แพทย์จะให้ยาแก่ผู้ป่วย} \entext{(The physician will give the patient medication)} &
\cmark & \cmark & \cmark \\
17 &
\thaitext{ แพทย์แนะนำให้ผู้ป่วยสังเกตว่า ต่อมทอนซิลบวมหรือมีหนองหรือไม่เมื่อกลับบ้าน } \entext{(The physician advises the patient to check whether the tonsils are swollen or have pus after returning home)} &
\xmark & \cmark & \xmark \\
18 &
\thaitext{ แพทย์อธิบายว่าในวันแรกของอาการ จะแยกได้ยากว่าเป็นการติดเชื้อแบคทีเรียหรือไวรัส } \entext{(The physician explains that on the first day of symptoms, it is hard to tell a bacterial from a viral infection)} &
\xmark & \xmark & \xmark \\
19 &
\thaitext{ แพทย์ประเมินจากอาการว่า ค่อนข้างเป็นการติดเชื้อไวรัส } \entext{(The physician assesses from the symptoms that it is likely a viral infection)} &
\cmark & \cmark & \cmark \\
20 &
\thaitext{ แพทย์ตรวจคลำต่อมน้ำเหลืองบริเวณคอ แล้วไม่พบว่ามีต่อมน้ำเหลืองโต } \entext{(The physician palpates the neck lymph nodes and finds no enlarged lymph nodes)} &
\cmark & \cmark & \cmark \\
21 &
\thaitext{ แพทย์อธิบายว่าการที่ต่อมน้ำเหลืองไม่โต มีแนวโน้มว่าเป็นการติดเชื้อไวรัสได้มากกว่า } \entext{(The physician explains that non-enlarged lymph nodes make a viral infection more likely)} &
\xmark & \xmark & \xmark \\
22 &
\thaitext{ แพทย์แนะนำให้กินยารักษาตามอาการ สำหรับการติดเชื้อไวรัส } \entext{(The physician advises symptomatic medication for the viral infection)} &
\cmark & \cmark & \cmark \\
23 &
\thaitext{ แพทย์แนะนำให้พักผ่อนเยอะ ๆ สำหรับการติดเชื้อไวรัส } \entext{(The physician advises plenty of rest for the viral infection)} &
\cmark & \cmark & \cmark \\
24 &
\thaitext{แพทย์เสนอใบลาให้ผู้ป่วย} \entext{(The physician offers the patient a leave note)} &
\xmark & \xmark & \cmark \\
25 &
\thaitext{ ผู้ป่วยต้องการใบลา 1 วันสำหรับวันพรุ่งนี้ } \entext{(The patient wants a 1-day leave note for tomorrow)} &
\xmark & \xmark & \xmark \\
26 &
\thaitext{แพทย์จะออกใบลาให้ผู้ป่วย} \entext{(The physician will issue a leave note for the patient)} &
\xmark & \cmark & \cmark \\
27 &
\thaitext{ผู้ป่วยไม่มีข้อสงสัยสอบถามเพิ่มเติม} \entext{(The patient has no further questions)} &
\xmark & \xmark & \xmark \\
28 &
\thaitext{แพทย์แนะนำให้ผู้ป่วยลองกินยา} \entext{(The physician advises the patient to try taking the medication)} &
\xmark & \cmark & \cmark \\
\midrule
\multicolumn{2}{r}{\small\textbf{Total entailed claims}} &
\textbf{17/28} & \textbf{22/28} & \textbf{21/28} \\
\multicolumn{2}{r}{\small\textbf{ClaimEntailment score}} &
\textbf{60.7} & \textbf{78.6} & \textbf{75.0} \\
\bottomrule
\end{tabularx}
\caption{
    Claim-level entailment comparison among ClinicEval, Ours trained on real encounters, and Ours
    trained on synthetic encounters. \cmark{} indicates that the reference claim is entailed by the
    generated SOAP note under the ClaimEntailment metric, whereas \xmark{} indicates that it is not
    fully entailed. Scores are reported as percentages.
}
\label{tab:qualitative_grpo_claim_entailment}
\end{table*}

\clearpage

\makeatletter\global\@colroom\@colht \global\vsize\@colht\makeatother
\section{Ethics \& Data Governance Statement}
\label{app:ethics}

\paragraph{ICOPD} consists of consented outpatient encounters collected exclusively for internal AI research. The dataset contains real physician--patient conversations and is used only for model development and evaluation within our institution. It is not publicly released or distributed to external researchers.

\paragraph{S-ICOPD} is a fully synthetic dataset generated from safe, de-identified clinical information derived from the ICOPD training data. It does not contain original physician--patient conversations or audio recordings. We will release S-ICOPD under a controlled-access, research-only data license to support reproducible research while minimizing the disclosure of real patient interactions.

\paragraph{ThaiClinicBench} was constructed from clinical encounters collected with signed informed consent from all participating patients and physicians. Participants were informed that their clinical dialogue data would be de-identified and used for AI research and dataset development. Participation was voluntary, and the collected data were used only for the purposes described in the consent materials.

To mitigate the associated privacy risks, all transcripts and reference notes underwent manual de-identification before model development or analysis. Direct identifiers and potentially identifying contextual information, based on the HIPAA Safe Harbor~\cite{hhs_hipaa_deidentification}, such as names, contact information, locations, dates, institutional identifiers, and other encounter-specific details, were removed or replaced with synthetic placeholders. Audio recordings are not included in the released dataset. The de-identification procedure was designed with reference to Thailand's Personal Data Protection Act~\cite{thailand_pdpa_2019} and the HIPAA Safe Harbor framework~\cite{hhs_hipaa_deidentification}.

ThaiClinicBench will be distributed under a controlled-access, research-only data license. Users must agree not to attempt to identify participants, link the data with external information for identification purposes, redistribute the original dataset, or use it for clinical decision-making. The dataset is intended solely for research on clinical language processing and medical-scribe systems. Requests concerning privacy or removal of released data may be directed to the ThaiClinicBench authors.

\end{document}